\documentclass[runningheads]{llncs}

\usepackage{eccv}

\usepackage{eccvabbrv}

\usepackage{graphicx}
\usepackage{booktabs}

\usepackage[accsupp]{axessibility}  

\usepackage{multirow}
\usepackage[table]{xcolor}
\usepackage{enumitem}
\usepackage{wrapfig}
\usepackage{amssymb}
\usepackage{inconsolata}
\definecolor{fftgray}{RGB}{235,235,235}
\definecolor{oursblue}{RGB}{220,235,250}
\definecolor{lorared}{RGB}{250,225,225}
\definecolor{sfgreen}{RGB}{225,245,225}
\usepackage{calc}
\newlength\myheight
\newlength\mydepth
\settototalheight\myheight{Xygp}
\usepackage{graphicx}
\usepackage{booktabs}
\usepackage{multirow}
\usepackage[table]{xcolor}
\definecolor{fftgray}{RGB}{235,235,235}
\definecolor{oursblue}{RGB}{220,235,250}
\definecolor{lorared}{RGB}{250,225,225}
\definecolor{sfgreen}{RGB}{225,245,225}
\usepackage{colortbl}
\usepackage{pgfmath}

\definecolor{rankone}{RGB}{255,215,0}
\definecolor{ranktwo}{RGB}{255,235,140}
\definecolor{rankthree}{RGB}{255,245,200}
\definecolor{rankfour}{RGB}{250,250,235}

\newcommand{\first}[1]{\cellcolor{rankone}\textbf{#1}}
\newcommand{\second}[1]{\cellcolor{ranktwo}#1}
\newcommand{\third}[1]{\cellcolor{rankthree}#1}

\usepackage{listings}

\DeclareRobustCommand{\taskNatural}{\raisebox{0.5pt}{\tikz{\fill[natural] (0cm,0cm) circle (.5ex);}}\,\textsc{natural}}
\DeclareRobustCommand{\taskSpecialized}{\raisebox{0.5pt}{\tikz{\fill[specialized] (0,0) circle (.5ex);}}\,\textsc{specialized}}
\DeclareRobustCommand{\taskStructured}{\raisebox{0.5pt}{\tikz{\fill[structured] (0,0) circle (.5ex);}}\,\textsc{structured}}

\definecolor{natural}{rgb}{0.7137,0.3333,0.3333}
\definecolor{specialized}{rgb}{0.4118,0.6431,0.4314}
\definecolor{structured}{rgb}{0.3254,0.4431,0.6666}
\usepackage[accsupp]{axessibility}

\usepackage{hyperref}

\usepackage{orcidlink}

\begin{document}

\title{LoCA: Spatially-Aware Low-Rank Convolutional Adaptation of Vision Foundation Models}

\titlerunning{LoCA}

\newcommand{\equalcontrib}{\textsuperscript{\scriptsize *}}
\newcommand{\corrauthor}{\textsuperscript{\scriptsize \textdagger}}
\author{
Sojung An\inst{1}\equalcontrib,\,
Junha Lee\inst{2,3}\equalcontrib,\,
Sujeong You\inst{2},\,
Nam Ik Cho\inst{3},\,
Donghyun Kim\inst{1}\corrauthor
}

\authorrunning{S.~An et al.}
\institute{
Korea University, Seoul, Republic of Korea
\and
Korea Institute of Industrial Technology, Ansan, Republic of Korea
\and
Seoul National University, Seoul, Republic of Korea\\[0.5em]
$^{*}$Equal contribution
$^{\dagger}$Corresponding author: \email{d\_kim@korea.ac.kr}\\
\raisebox{-\mydepth}{\includegraphics[height=1.\myheight]{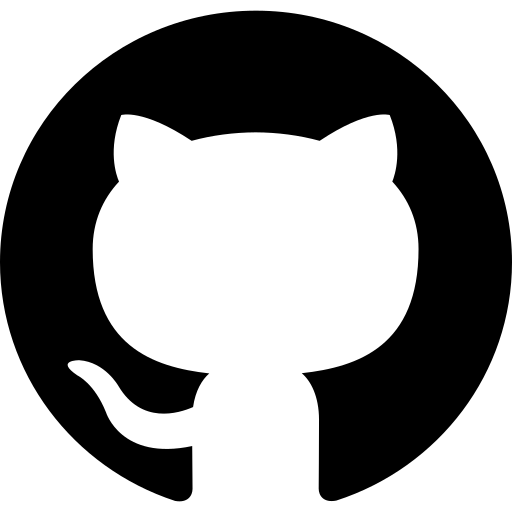}}
\textbf{\url{https://github.com/ssojungan/loca}}
}

\maketitle

\begin{abstract}
Pre-trained Vision Foundation Models (VFMs) provide strong visual representations for diverse downstream tasks.
The key challenge of VFM adaptation stems from the prohibitive costs of full fine-tuning and catastrophic forgetting.
To address this, Low-Rank Adaptation (LoRA) has emerged as the prevailing paradigm for Parameter-Efficient Fine-Tuning (PEFT).
However, LoRA is typically designed for transformer self-attention layers parameterized by 2D matrices.
Since convolutional kernels inherently couple spatial and channel information within a 4D tensor, forcing them into a monolithic 2D matrix disrupts the inherent spatial topology.
In this paper, we propose Low-Rank Convolutional Adaptation (LoCA), a convolution-aware PEFT framework that addresses spatial-channel entanglement by decoupling channel and spatial adaptation.
LoCA introduces a low-rank channel adaptation for dense cross-channel mixing and refines spatial bases extracted from pre-trained kernels via Singular Value Decomposition (SVD).
Experimental results show that LoCA preserves pre-trained spatial priors and achieves competitive or state-of-the-art performance across fine-grained classification, domain-generalized semantic segmentation, and generative benchmarks.
\keywords{Parameter-Efficient Fine-Tuning \and Low-Rank Adaptation \and Spatial Inductive Bias \and  Convolutional Networks}
\end{abstract}

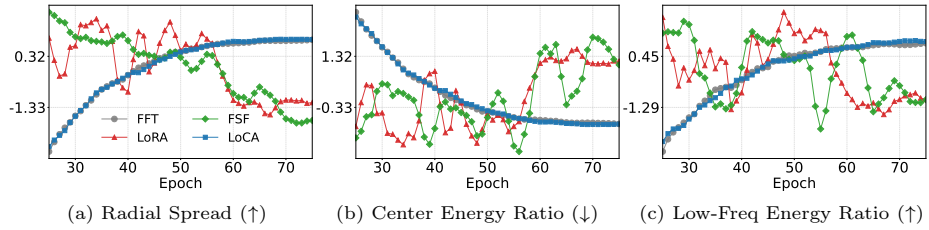
\begin{figure}[t]
    \centering
    \setlength{\fboxsep}{0pt}
    \begin{tikzpicture}
        \node[anchor=south west,inner sep=0] (img) at (0,0)
        {\includegraphics[width=\linewidth]{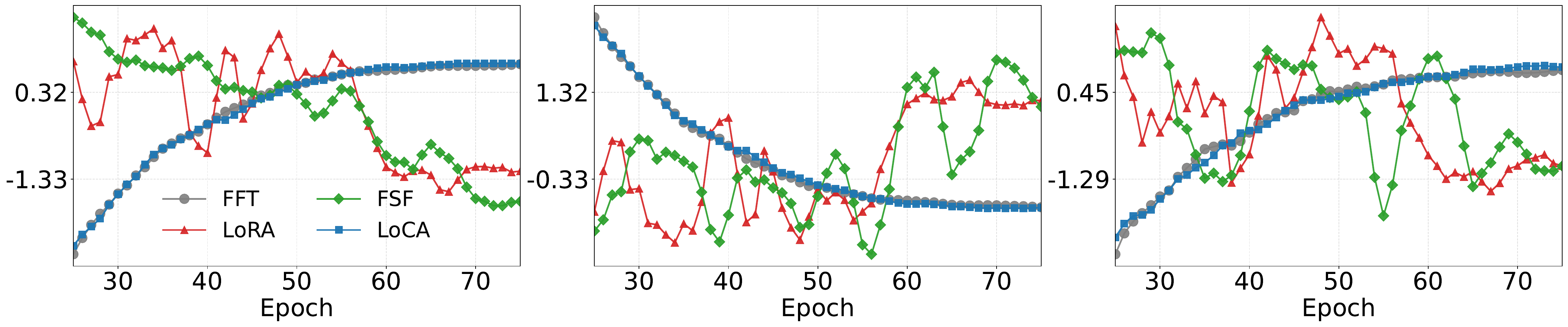}};
        \begin{scope}[x={(img.south east)},y={(img.north west)}]
            \node at (0.175,-0.1) {\scriptsize (a) Radial Spread ($\uparrow$)};
            \node at (0.5,-0.1) {\scriptsize (b) Center Energy Ratio ($\downarrow$)};
            \node at (0.84,-0.1) {\scriptsize (c) Low-Freq Energy Ratio ($\uparrow$)};
        \end{scope}
    \end{tikzpicture}
    \caption{The training dynamics of Effective Receptive Field (ERF) across convolutional layers ($dwconv$) for \colorbox{fftgray}{\strut FFT}, \colorbox{lorared}{\strut LoRA}, \colorbox{sfgreen}{\strut FSF}, and \colorbox{oursblue}{\strut LoCA}. (a) Radial Spread: Average distance of gradients from the map center. (b) Center Energy Ratio: Gradient ratio in the $3 \times 3$ center. (c) Low-Freq Energy: Fourier low-frequency power ratio. LoRA and FSF exhibit transient ERF growth followed by a reconvergence toward localized patterns, which limit sustained global context. Spatial reconvergence necessitates a structure-preserving design to achieve an expansive receptive field comparable to FFT.}
    \label{fig:motivation}
\end{figure}

\section{Introduction}
Vision Foundation Models (VFMs) enable a wide array of general vision tasks~\cite{convnext, vit, dinov3}.
These models learn rich multi-scale representations from large-scale data and transfer effectively to downstream problems through fine-tuning or frozen feature extraction~\cite{kim2022broad, li2022exploring}.
However, full fine-tuning of large-scale VFMs not only incurs prohibitive computational costs but also leads to catastrophic forgetting of pre-trained knowledge.
Parameter-Efficient Fine-Tuning (PEFT) addresses this by optimizing only a small subset of parameters while freezing the original weights~\cite{xu2023parameter, han2024parameter}.
Among these, Low-Rank Adaptation (LoRA) has emerged as the de facto standard, offering high representational capacity with minimal trainable parameters~\cite{lora, zhang2023adalora, sora, yun2025soma}.

While LoRA is typically designed for linear projections in transformers, convolutional operators in VFM backbones remain relatively underexplored.
This gap is critical because convolutional operators remain fundamental across modern VFM backbones.
ConvNeXt is a modern convolutional backbone that emphasizes convolution as a sliding-window, weight-sharing strategy that encodes spatial inductive biases~\cite{woo2023convnext}.
Self-attention in ViT~\cite{vit} enables global token-to-token interactions but provides limited built-in spatial inductive bias.
This prompts hybrid designs that incorporate convolutional components to encode spatial priors~\cite{chen2023vision}.
Mamba-based State Space Model (SSM) vision backbones have gained recent attention~\cite{gu2024mamba, mambavision}.
Some architectures retain convolutional blocks in early high-resolution stages for local feature extraction. 
In addition to visual perception, convolutional networks are widely used in generative models such as Stable Diffusion~\cite{hrnet} in the U-Net backbone.
Given that convolutional operators remain fundamental across modern VFM backbones, extending LoRA beyond Transformer linear layers to convolutional layers becomes an important problem in PEFT.

However, directly applying LoRA’s low-rank updates to convolution layers is suboptimal, as they operate over spatial regions. Naive LoRA flattens convolutional kernels (4D tensor) into a two-dimensional matrix for low-rank updates~\cite{ding2024lora}.
Flattening the convolutional kernel collapses the inherent spatial topology by enforcing cross-channel mixing within a single low-rank parameterization.
This structural mismatch limits the preservation of spatial priors, including locality and directionality, and reduces adaptation gains reported in prior work~\cite{zhong2024convolution, chen2025large}.
Recent filter subspace approaches address spatial–channel entanglement by decomposing convolutional filters into spatial bases and channel coefficients~\cite{chen2025large}.
Sparse coding approximates pre-trained kernels within this decomposed subspace. 
Such approximation inevitably modifies pre-trained representations prior to fine-tuning.
Freezing cross-channel mixing coefficients limits the adaptation of inter-channel correlations in new domains. 

\begin{figure}[t]
    \centering
    \begin{subfigure}[b]{0.55\linewidth}
        \centering
        \includegraphics[width=\linewidth,clip,trim=0 0 0 0]{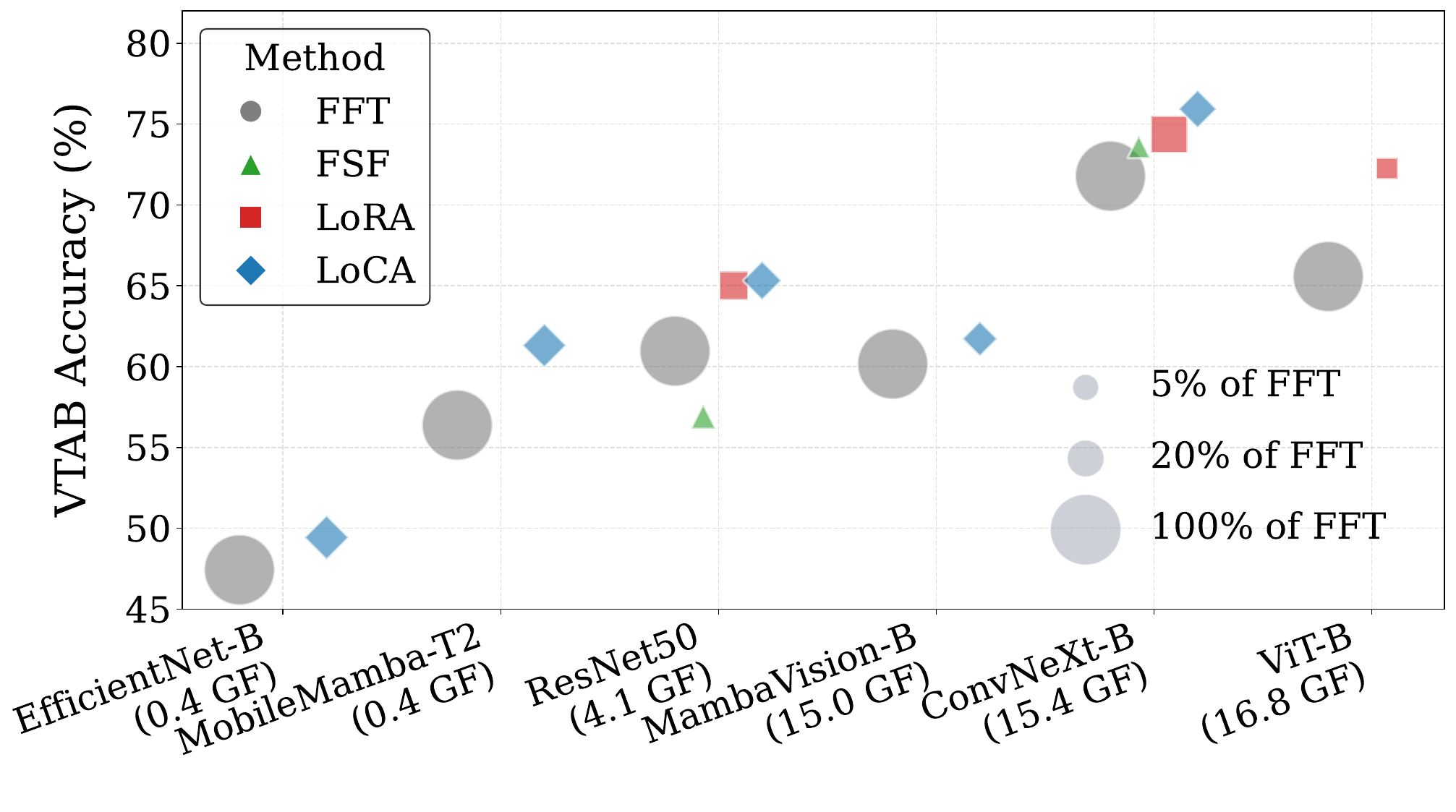}
        \vspace{-1.5em}
        \caption{Parameter-efficient scaling behavior. The x-axis is sorted by GFLOPs.}
        \label{fig:result}
    \end{subfigure}\hfill
    \begin{subfigure}[b]{0.42\linewidth}
        \centering
        \includegraphics[width=\linewidth,clip,trim=0 120pt 0 120pt]{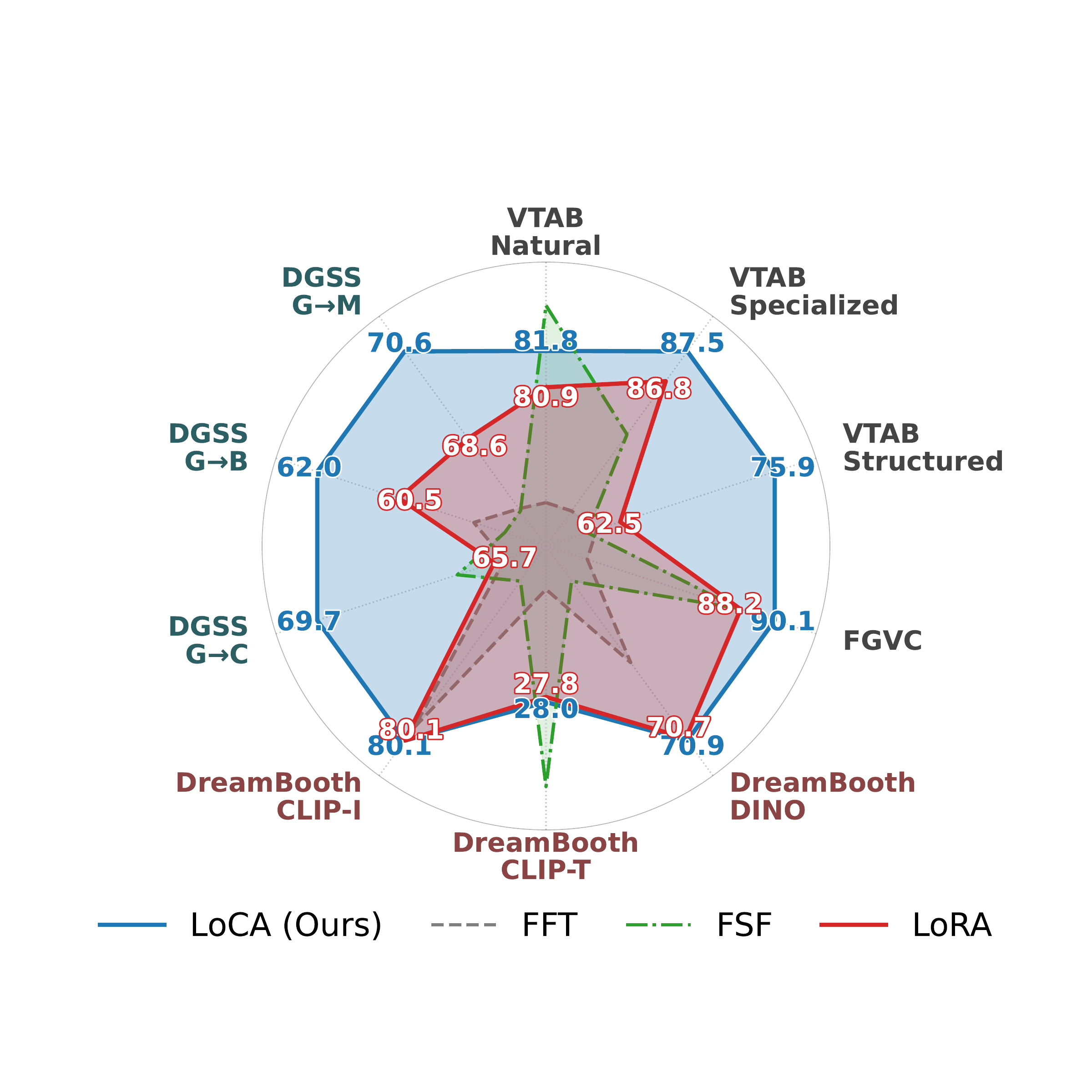}
        \vspace{-1.5em}
        \caption{Generalization performance of PEFT methods across diverse vision benchmarks}
        \label{fig:LoCA}
    \end{subfigure}
    \vspace{1.5em}
    \begin{subfigure}[b]{0.95\linewidth}
        \centering
        \includegraphics[width=\linewidth,clip,trim=0 0 0 0]{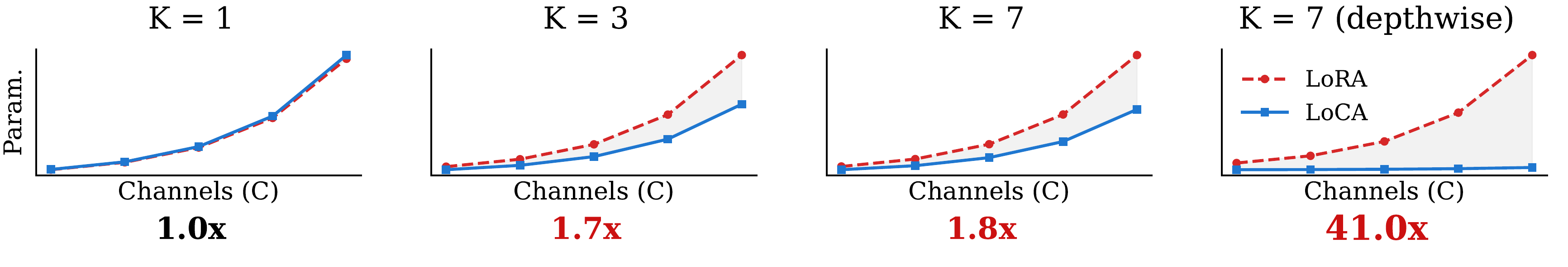}
        \vspace{-1.5em}
        \caption{Trainable parameter comparison across convolution kernel sizes (K). Detailed analysis is provided in Appendix A.}
        \label{fig:param_comparison}
    \end{subfigure}\hfill
    \vspace{-1.5em}
    \caption{Performance comparison of PEFT-based methods on downstream tasks}
    \label{fig:peft_LoCA}
\end{figure}

As shown in \cref{fig:motivation}, we analyze the Effective Receptive Field (ERF) \cite{he2025mobilemamba} and low-frequency retention during fine-tuning.
Higher Radial Spread with lower Center Energy Ratio indicates outward gradient propagation and receptive field growth, while higher Low-Freq Energy reflects stronger retention of informative low-frequency structures.
FFT and LoCA show consistent increases in spatial coverage and low-frequency energy throughout training.
In contrast, LoRA \cite{lora} and Filter Subspace Fine-Tuning (FSF) \cite{chen2025large} exhibit only transient spatial expansion, followed by reconvergence toward localized update patterns and unstable frequency behavior.
These observations suggest that preserving convolutional spatial structure is crucial for maintaining broad receptive fields, motivating our spatial–channel disentanglement approach with joint adaptation.

To this end, we propose Low-Rank Convolutional Adaptation (LoCA), a structured low-rank reparameterization that adapts convolutional layers by decoupling channel mixing from spatial basis refinement.
We first introduce a low-rank channel adaptation process that captures dense cross-channel mixing while mitigating spatial–channel entanglement.
Such channel adaptation prevents the topological collapse induced by naive kernel flattening and the structural inaccuracies of conventional weight decomposition.
We then design a spatial adaptation mechanism that preserves pre-trained priors through structural bases derived from Singular Value Decomposition (SVD).
Additionally, we introduce a hierarchical rank scheduling for convolutional foundation models.
LoCA preserves the spatial inductive bias of pre-trained representations and achieves robust performance across diverse downstream tasks.

We summarize the contributions of this work as follows:
\begin{itemize}[topsep=0.1pt, itemsep=0.pt]
    \item We propose the LoCA framework to address spatial-channel entanglement by decoupling channel and spatial adaptation.
    \item We introduce SVD-based spatial basis refinement to preserve pre-trained spatial inductive biases effectively.
    \item We propose a hierarchical rank scheduling tailored to convolutional foundation models.
    \item Extensive experiments demonstrate that LoCA achieves competitive or state-of-the-art performance across fine-grained classification, domain-generalized segmentation, and generative benchmarks (see \cref{fig:peft_LoCA}).
\end{itemize}

\section{Related Work}
\subsection{Parameter-Efficient Fine-Tuning}
PEFT has emerged as a practical paradigm for adapting large-scale VFMs to downstream tasks by updating only a small subset of parameters while freezing pre-trained backbones~\cite{vpt, lora}. 
Existing approaches include adapter-based methods~\cite{mlp, convadapter}, which insert lightweight trainable modules into network blocks; prompt-based methods~\cite{vpt, li2021prefix}, which add learnable tokens to the input sequence; selective fine-tuning methods~\cite{bias, guo2021parameter}, which update specific components such as bias terms; and reparameterization-based methods~\cite{lora, zhang2023adalora, sora, yun2025soma}, which optimize implicit low-rank structures for seamless merging into the original weights at inference.

Most PEFT techniques are formulated for Transformer architectures that operate on token sequences with multi-head self-attention~\cite{prottasha2025peft}. 
Extending these paradigms to vision tasks such as detection and segmentation remains challenging because pixel-level prediction relies on spatial inductive biases and multi-scale hierarchies. 

\subsection{Parameter-Efficient Fine-Tuning for Convolutional Layers}
Convolution preserves 2D image structure by leveraging inherent spatial inductive biases~\cite{convnext, luo2016understanding}.
Early convolution-specific PEFT methods such as Conv-Adapter~\cite{convadapter} introduce trainable modules into convolutional blocks and increase inference-time computation.
Flattening-based extensions convert 2D convolutional kernels parameterized as 4D tensors into 2D matrices to reuse linear LoRA formulations~\cite{ding2024lora}. 
This dimensional collapse entangles spatial topology with cross-channel mixing, compromising locality and weight sharing~\cite{zhong2024convolution, chen2025large}.

Filter subspace methods constrain updates by decomposing pre-trained kernels into spatial atoms and mixing coefficients~\cite{chen2025large}. 
This decomposition reconstructs the pre-trained weights only approximately while introducing accumulation error.
Reconstructing weights from these decomposed atoms can also restrict cross-channel flexibility by freezing mixing coefficients.
These limitations motivate a convolution-aware PEFT that preserves the original weights while enabling spatially structured low-rank updates.

\subsection{Low-Rank Adaptation and Singular Value Decomposition}
LoRA represents weight adaptation using low-rank factors.
To maximize parameter efficiency, subsequent methods have explored rank adaptation.
For example, AdaLoRA dynamically adjusts the rank budget across different layers based on importance scores~\cite{zhang2023adalora}.
Other works leverage SVD to replace random initialization with informed low-rank initialization.
PiSSA~\cite{pissa}, SoMA~\cite{yun2025soma}, and SoRA~\cite{sora} decompose pre-trained weights and use principal or minor components to initialize low-rank factors, improving convergence and knowledge retention.
These works suggest that singular components capture reusable structure in pre-trained weights.
Building upon this idea, we employ SVD to extract spatial bases from convolutional kernels.

\section{Preliminaries}
\noindent\textbf{LoRA.} LoRA freezes pre-trained weights and approximates updates using low-rank matrices.
Motivated by the hypothesis that weight changes during model adaptation possess a low intrinsic rank, LoRA parameterizes the incremental update via the product of two low-rank matrices \cite{lora}. 
For a pre-trained weight matrix $W_0 \in \mathbb{R}^{d_{out} \times d_{in}}$, where $d_{in}$ and $d_{out}$ denote the input and output dimensions respectively, LoRA decomposes the update $\Delta W \in \mathbb{R}^{d_{out} \times d_{in}}$ into $BA$, where $B \in \mathbb{R}^{d_{out} \times r}$ and $A \in \mathbb{R}^{r \times d_{in}}$ are low-rank matrices with rank $r \ll \min(d_{out}, d_{in})$. and $\alpha$ is a constant scaling factor. Consequently, the fine-tuned weight $W'$ is formulated as:
\begin{equation}
    W' = W_0 + \Delta W = W_0 + \frac{\alpha}{r} B A
    \label{eq:lora_linear}
\end{equation}
where $W_0$ remains frozen during training and $\alpha$ denotes a constant scaling factor.
One factor is zero-initialized, yielding $\Delta W=0$ at initialization.

A 2D convolutional layer is parameterized by a 4D weight tensor $W_0 \in \mathbb{R}^{C_{out} \times C_{in} \times k_h \times k_w}$ with $C_{out}$ output channels, $C_{in}$ input channels, and spatial kernel size $k_h \times k_w$.
LoRA is formulated for matrix multiplication, so a naive extension to convolution reshapes the kernel tensor into a matrix.
This reshaping merges the spatial dimensions into the input-channel axis and produces the flattened weight $W_{0}^{\flat} \in \mathbb{R}^{C_{out} \times (C_{in}k_h k_w)}$.
The low-rank update is computed in the flattened space by applying Eq.~\eqref{eq:lora_linear} to $W_0^{\flat}$:
\begin{equation}
    W'^{\flat} = W_{0}^{\flat} + \frac{\alpha}{r} B A,
    \label{eq:lora_naive}
\end{equation}
where $B \in \mathbb{R}^{C_{\mathrm{out}} \times r}$ and $A \in \mathbb{R}^{r \times (C_{in}k_h k_w)}$.

\noindent\textbf{Filter Subspace Adaptation.} 
Filter Subspace Fine-tuning (FSF) was proposed to represent each convolution filter as a linear combination of spatial elements referred to as filter atoms.
Formally, Chen et al.~\cite{chen2025large} decompose a pre-trained convolutional layer $W_0 \in \mathbb{R}^{C_{out} \times C_{in} \times k_h \times k_w}$ into a filter atom layer $\mathbf{D} \in \mathbb{R}^{m \times k_h \times k_w}$ and an atom coefficient layer $\boldsymbol{\alpha} \in \mathbb{R}^{C_{out} \times C_{in} \times m}$:
\begin{equation}
    W_0 = \boldsymbol{\alpha} \times \mathbf{D}.
\end{equation}
This indicates that each filter slice $W_0^{i,j} \in \mathbb{R}^{k_h \times k_w}$ is constructed by a linear combination of the filter atoms: $W_0^{i,j} = \sum_{l=1}^{m} \boldsymbol{\alpha}^{i,j,l} \mathbf{d}_{l}$. 
$\mathbf{D} = \{\mathbf{d}_l\}_{l=1}^m$ denotes a set of $m$ filter atoms for spatial convolution and $\boldsymbol{\alpha}$ controls spatially invariant cross-channel mixing.
Sparse coding initializes these components by minimizing reconstruction error on the pre-trained weights.
Each filter atom $\mathbf{d}_l$ can be recursively decomposed into an overcomplete atom set $\mathbf{D}_1 \in \mathbb{R}^{(m \cdot m_1) \times k_h \times k_w}$ using intra-channel mixing coefficients $\boldsymbol{\beta} \in \mathbb{R}^{(C_{in} \cdot m) \times m_1}$.
FSF updates only the spatial atoms ($\mathbf{D}$ or $\mathbf{D}_1$) while freezing $\boldsymbol{\alpha}$ to retain pre-trained generalization. 

\section{Low-Rank Convolutional Adaptation}
\label{sec:loca}
This section introduces LoCA as a framework that preserves spatial inductive bias during convolutional layer adaptation.
LoCA decouples the adaptation into low-rank channel adaptation (\cref{sec:channel}) and spatial basis refinement (\cref{sec:spatial}).
The low-rank channel adaptation isolates dense cross-channel mixing to resolve spatial–channel entanglement, while spatial basis refinement optimizes SVD-derived structural bases to preserve the inherent spatial topology.
The independent channel and spatial paths are then composed to form the decoupled LoCA design (\cref{sec:integration}).
Finally, we introduce hierarchical rank scheduling for convolutional vision backbones (\cref{sec:hierarchy}).

\begin{figure}[t]
     \centering
     \includegraphics[width=\linewidth]{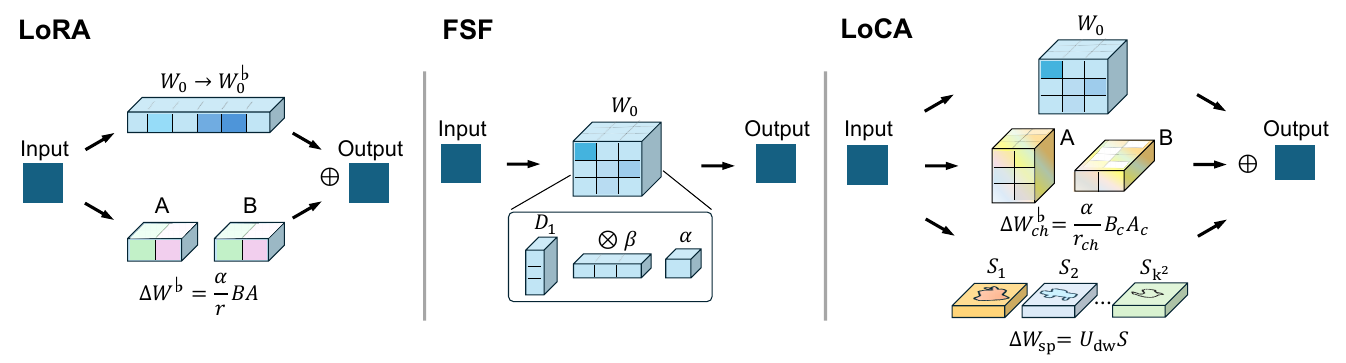}
     \caption{Convolutional kernel adaptation architectures. (a) LoRA: Decomposes weights into a frozen $W_0$ and a low-rank update $\Delta W = BA$. (b) FSF: Decomposes $W_0$ into spatial atom bases ($D_1$), intra-channel mixing components ($\beta$), and cross-channel atom coefficients ($\alpha$). (c) LoCA (ours): Learns only $\Delta W$ while freezing $W_0$, integrating spatial information via learned basis $\mathcal{S}$ for stable channel–spatial structural adaptation.}
     \label{fig:method}
\end{figure}

\subsection{Low-Rank Channel Adaptation}
\label{sec:channel}
To explicitly resolve the spatial-channel entanglement, we establish a dedicated low-rank channel adaptation mechanism that isolates dense cross-channel dependencies from spatial topology.
Building upon the flattened formulation in Eq.~\eqref{eq:lora_naive}, we define the low-rank channel adaptation term.
Let $r_{\mathrm{ch}}$ denote the channel rank and $\alpha$ the scaling factor. The update is defined as:
\begin{equation}
    \Delta W_{\mathrm{ch}}^{\flat} = \frac{\alpha}{r_{\mathrm{ch}}} B_{c} A_{c},
\end{equation}
where $B_c \in \mathbb{R}^{C_{\mathrm{out}} \times r_{\mathrm{ch}}}$ and $A_c \in \mathbb{R}^{r_{\mathrm{ch}} \times (C_{\mathrm{in}} k_h k_w)}$.
We then reshape $\Delta W_{\mathrm{ch}}^{\flat}$ back to its original 4D spatial structure, denoted as $\Delta W_{\mathrm{ch}} \in \mathbb{R}^{C_{\mathrm{out}} \times C_{\mathrm{in}} \times k_h \times k_w}$.
To guarantee functional equivalence to the pre-trained model at initialization, $A_c$ is initialized using a Kaiming uniform distribution, and $B_c$ is initialized to zero.
This zero-initialization ensures $\Delta W_{\mathrm{ch}} = 0$ at the start of training, preserving the original pre-trained representations without requiring kernel replacement.

\subsection{SVD-based Spatial Basis Refinement}
\label{sec:spatial}
Naive flattening collapses the $k_h \times k_w$ spatial structure, making it difficult to isolate spatial modulation from dense channel transformations.
We instead parameterize spatial adaptation with a compact set of pre-trained spatial bases.
In practice, we reshape each pre-trained kernel slice into a length-$k_h k_w$ vector and apply zero-mean and unit-variance standardization to obtain $W_{\mathrm{norm}} \in \mathbb{R}^{C_{\mathrm{out}} C_{\mathrm{in}} \times k_h k_w}$. We then form a spatial covariance matrix $C_{\mathrm{sp}} = W_{\mathrm{norm}}^{\top} W_{\mathrm{norm}} \in \mathbb{R}^{k_h k_w \times k_h k_w}$.
Specifically, we compute the SVD of the spatial covariance induced by the pre-trained kernel, set the spatial rank as $r_{\mathrm{sp}} = k_h k_w$, and obtain an initial basis tensor $\mathcal{S} \in \mathbb{R}^{r_{\mathrm{sp}} \times k_h \times k_w}$.
The deterministic basis $\mathcal{S}$ is initialized from this SVD and treated as a learnable parameter for refinement.
Each slice $\mathcal{S}_m \in \mathbb{R}^{k_h \times k_w}$ corresponds to the $m$-th principal spatial pattern, such as edges or textures.
Channel-specific coefficients are learned with $U_{\mathrm{dw}} \in \mathbb{R}^{D \times r_{\mathrm{sp}}}$ where $D = \min(C_{\mathrm{out}}, C_{\mathrm{in}})$.
The spatial update $\Delta W_{\mathrm{sp}}^{\mathrm{diag}}[i] \in \mathbb{R}^{k_h \times k_w}$ for channel $i$ is defined by a basis expansion:
\begin{equation}
    \Delta W_{\mathrm{sp}}^{\mathrm{diag}}[i] = \sum_{m=1}^{r_{\mathrm{sp}}} U_{\mathrm{dw}}[i, m] \cdot \mathcal{S}_m.
\end{equation}
The spatial tensor is defined on the depthwise diagonal with the Kronecker delta $\delta_{ij}$ (i.e., $\delta_{ij}=1$ if $i=j$ and $0$ otherwise).
Since $i=j$ implicitly guarantees $i \le \min(C_{\mathrm{out}}, C_{\mathrm{in}}) = D$, the index is safely bounded:
\begin{equation}
    \Delta W_{\mathrm{sp}}[i,j,:,:] = \delta_{ij}\,\Delta W_{\mathrm{sp}}^{\mathrm{diag}}[i].
\end{equation}
Diagonal parameterization separates spatial refinement from cross-channel mixing.

\subsection{Channel-Spatial Composition}
\label{sec:integration}
We compose the independent channel and spatial paths to synthesize our decoupled design into a unified adaptation framework without distorting pre-trained knowledge.
The combined update tensor $\Delta W_{cs} \in \mathbb{R}^{C_{\mathrm{out}} \times C_{\mathrm{in}} \times k_h \times k_w}$ adds the spatial update on the depthwise diagonal and uses $\Delta W_{\mathrm{ch}}$ for cross-channel mixing:
\begin{equation}
    \Delta W_{cs}[i, j, :, :] = \Delta W_{\mathrm{ch}}[i, j, :, :] + \delta_{ij} \Delta W_{\mathrm{sp}}^{\mathrm{diag}}[i]
\end{equation}
Initializing $U_{\mathrm{dw}}$ and $B_c$ to zero ensures $\Delta W_{cs} = 0$, thereby preserving exact functional equivalence to the frozen pre-trained model.
The final adapted weight $W'$ is defined as follows:
\begin{equation}
    W' = W_0 + \Delta W_{cs}.
    \label{eq:loca_final}
\end{equation}

\subsection{Hierarchical Rank Scheduling}
\label{sec:hierarchy}
Convolution-based VFMs adopt a hierarchical architecture that encodes features through progressively increasing channel dimensions across stages \cite{hrnet, woo2023convnext}. 
Early stages extract local features with narrower channels, while deeper stages encode global semantics with wider channels. 
A fixed rank applied uniformly across stages limits the ability to capture this hierarchical diversity. 
To address this, we introduce hierarchical rank scheduling to determine the channel rank $r_{\mathrm{ch}}^{(s)}$ based on the stage-specific width $C_{\mathrm{out}}^{(s)}$. 
We compute a base rank as $\lfloor R \cdot \frac{C_{\mathrm{out}}^{(s)}}{\sum_{i} C_{\mathrm{out}}^{(i)}} \rfloor$, where $R$ is the global rank budget.
By aligning adaptation capacity with layer width, this strategy ensures that the narrower stem receives smaller ranks while deeper layers are allocated larger ranks to model complex semantic features. 
While the channel rank is scaled across stages, the spatial rank $r_{\mathrm{sp}}$ (as defined in \cref{sec:spatial}) remains fixed to the localized spatial kernel size ($k_h k_w$).
This strategy improves parameter efficiency without sacrificing adaptation performance. 
Empirical validation is provided in \cref{sec:ablation}.

\section{Experiment}
In this section, we evaluate LoCA's effectiveness through experiments across three distinct tasks: (1) fine-grained visual adaptation (\cref{sec:fg}) using the VTAB-1k and FGVC datasets; (2) generative generalization performance (\cref{sec:gen}) via the DreamBooth dataset; and (3) domain generalized semantic segmentation (DGSS).

\subsection{Fine-grained Adaptation on VTAB-1k and FGVC}
\label{sec:fg}

\noindent\textbf{Experimental Setup.}
Our evaluation utilizes the FGVC and VTAB-1k benchmarks. The FGVC comprises four fine-grained recognition tasks: CUB-200-2011 \cite{CUB_200_2011}, Stanford Dogs \cite{dataset2011novel}, Stanford Cars \cite{krause20133d}, and NABirds \cite{van2015building}. The VTAB-1k benchmark partitions downstream tasks into Natural, Specialized, and Structured semantic domains.

\noindent\textbf{Results.}
We compare the proposed LoCA with existing PEFT methods on VTAB-1k and FGVC benchmarks.
As shown in \cref{tab:vtab}, we report LoCA results on ConvNeXt-B and ResNet-50.
On VTAB-1k, rank-16 (r16) generally achieves the highest accuracy, while ConvNeXt-B exhibits only marginal gains from increasing the rank.
The marginal gains on ConvNeXt-B stem from the smaller dimensionality of convolutional kernels compared with linear layers.
LoCA achieves the best VTAB-1k average accuracy with rank-4 on ConvNeXt-B and rank-16 on ResNet-50.
On ConvNeXt-B, LoCA reaches this performance with only 0.97 M trainable parameters.
In \cref{tab:fgvc}, LoCA with rank-16 outperforms the other PEFT methods on FGVC using both ConvNeXt-B and ResNet-50.
LoRA applies low-rank approximation after kernel flattening, which increases the number of additional parameters to 17.58 M on ConvNeXt-B while yielding lower accuracy.
Due to architectural differences between ConvNeXt (\texttt{Conv. Seq.}) and ResNet (\texttt{Res. Par.}), we use sequential CA insertion for ConvNeXt and parallel $k \times k$ CA adaptation for ResNet, following the best-performing placements reported in Conv-Adapter \cite{convadapter}.

\begin{table}[t]
\centering
\caption{Performance comparison on the VTAB-1k visual classification benchmark using ConvNeXt-B and ResNet-50 backbones. We compare LoCA (in blue) with full Fine-Tuning (FFT) (in gray), Linear Probing (LP), Partial \cite{partial}, MLP \cite{mlp}, Bias Tuning (Bias) \cite{bias}, Visual Prompt Tuning (VPT) \cite{vpt}, Conv-Adapter \cite{convadapter}, and Filter Subspace Fine-Tuning (FSF) \cite{chen2025large}. Param. represents the trainable parameters (M).}
\vspace{-1em}
\label{tab:vtab}
\resizebox{0.99\columnwidth}{!}{%
\begin{tabular}{@{}l|r|cccc|r|cccc@{}}
\toprule[1pt]
  \multicolumn{1}{c|}{\multirow{2}{*}{Tuning}} &
  \multicolumn{1}{c|}{\multirow{2}{*}{Param.}} &
  \multicolumn{4}{c|}{ConvNeXt-B} & 
  \multicolumn{1}{c|}{\multirow{2}{*}{Param.}} &
  \multicolumn{4}{c}{ResNet-50} \\ 
  \cmidrule(l){3-6} \cmidrule(l){8-11}
  & \multicolumn{1}{c|}{} &
  Natural &
  Specialized &
  Structured &
  Average &
  \multicolumn{1}{c|}{} &
  Natural &
  Specialized &
  Structured & 
  Average \\ \midrule
\# Tasks
& - & 7 & 4 & 8 & 19 & - & 7 & 4 & 8 & 19 \\ \midrule
    \rowcolor{fftgray} FFT & 87.62 & 80.52 & 87.54 & 63.85 & 74.98 & 23.61 & 65.58 & 82.0 & 52.32 & 63.45 \\
    \midrule
    LP & 1.68 & 74.48 & 81.50 & 34.76 & 59.23 & 0.48 & 63.75 & 77.60 & 30.96 & 52.89\\
    Partial-1 \cite{partial} & 4.72 & 73.76 & 81.64 & 39.55 & 61.01 & 2.10 & 64.34 & 78.64 & 45.78 & 59.51\\
    MLP-3 \cite{mlp} & 2.45 & 73.78 & 81.36 & 35.68 & 59.33 & 3.51 & 61.79 & 70.77 & 33.97 & 51.97\\
    Bias \cite{bias} & 1.76 & 69.07 & 72.81 & 25.29 & 51.42 & 0.49 & 63.51 & 77.22 & 33.39 & 53.85\\
    VPT \cite{vpt} & 1.75 & 78.48 & 83.00 & 44.64 & 65.18 & 0.49 & 66.25 & 77.32 & 37.52 & 56.09\\
    LoRA \cite{lora} & 17.32 & 80.89 & 86.80 & 62.46 & 74.36 & 1.90 & 65.06 & 82.53 & \textbf{56.21} & 65.01 \\
    CA \cite{convadapter} & 6.83 & 80.62 & 86.29 & 64.88 & 75.18 & 1.37 & 64.20 & 81.33 & 52.74 & 64.78 \\
    CoLoRA \cite{colora} & 4.57 & 76.1 & 83.1 & 58.2 & 70.0 & 1.40 & 66.6 & 82.6 & 51.9 & 63.8 \\
    FSF \cite{chen2025large} & 1.11 & \textbf{82.96} & 85.53 & 59.59 & 73.59 & 0.72 & 62.64 & 80.25 & 36.50 & 56.91 \\
    \midrule
    \rowcolor{oursblue} LoCA (r4) & 0.97 & 82.22 & \textbf{87.54} & 64.74 & \textbf{75.98} & 0.47 & 66.08 & 82.41 & 52.44 & 63.77 \\
    \rowcolor{oursblue} LoCA (r16) & 5.01  & 81.81 & 87.51 & \textbf{65.01} & 75.93 & 1.51 & \textbf{67.21} & \textbf{83.61} & 54.52 & \textbf{65.32} \\
\bottomrule[1pt]
\end{tabular}%
}
\end{table}

\begin{table}[t]
\centering
\begin{minipage}{0.48\columnwidth}
\centering
\caption{Average Top-1 accuracy ($\%$) on FGVC datasets}
\vspace{-1em}
\resizebox{\linewidth}{!}{%
\begin{tabular}{l|r|c|r|c}
\toprule[1pt]
\multirow{2}{*}{Tuning} & 
\multicolumn{2}{c|}{ConvNeXt-B} & 
\multicolumn{2}{c}{ResNet-50} \\ 
\cmidrule(l){2-3} \cmidrule(l){4-5}
& \# Param. & Average & \# Param. & Average \\ 
\midrule
\rowcolor{fftgray}
FFT & 87.87 & 79.73 & 24.14 & 75.73 \\
\midrule
LP & 0.31 & 77.55 & 0.62 & 45.39 \\
Bias \cite{bias} & 0.44 & 64.98 & 0.67 & 48.85  \\
LoRA \cite{lora} & 17.58 & 88.18 & 2.43 & 80.96 \\
CA \cite{convadapter} & 6.13 & 89.28 & 2.23 & 83.48 \\
CoLoRA \cite{colora} & 4.57 & 86.11 & 0.99 & 76.44 \\
FSF \cite{chen2025large} & 1.11 & 88.04 & 2.52 & 81.07 \\
\midrule
\rowcolor{oursblue}
LoCA (r16) & 3.70 & \textbf{90.06} & 1.94 & \textbf{83.67} \\
\bottomrule[1pt]
\end{tabular}%
}
\label{tab:fgvc}
\end{minipage}
\hfill
\begin{minipage}{0.48\columnwidth}
\centering
\caption{Quantitative comparison of subject alignment}
\vspace{-1em}
\resizebox{\linewidth}{!}{%
\begin{tabular}{lccc}
\toprule
Methods & DINO$\uparrow$ & CLIP-I$\uparrow$ & CLIP-T$\uparrow$ \\
\midrule
Pretrained & 0.320 & 0.643 & 0.267 \\
Real Images & 0.711 & 0.857 & -- \\
\midrule
Textual Inversion~\cite{gal2022textual} & 0.564 & 0.739 & 0.213 \\
\rowcolor{fftgray} DreamBooth~\cite{dreambooth} & 0.642 & 0.794 & 0.236 \\
LoRA~\cite{lora} & 0.637 & 0.792 & 0.239 \\
\quad $\hookrightarrow$ w/ Conv & 0.707 & 0.801 & 0.278 \\
FSF~\cite{chen2025large} & 0.572 & 0.715 & \textbf{0.313} \\
\midrule
\rowcolor{oursblue} LoCA (r16) & \textbf{0.709} & \textbf{0.801} & 0.280 \\
\bottomrule
\end{tabular}}
\label{tab:dreambooth}

\end{minipage}
\end{table}

\noindent\textbf{Generalization across Backbones.} The performance of LoCA is evaluated 
\begin{wrapfigure}[9]{r}{0.62\columnwidth}
    \centering
    \vspace{-2.4em}
    \includegraphics[width=\linewidth]{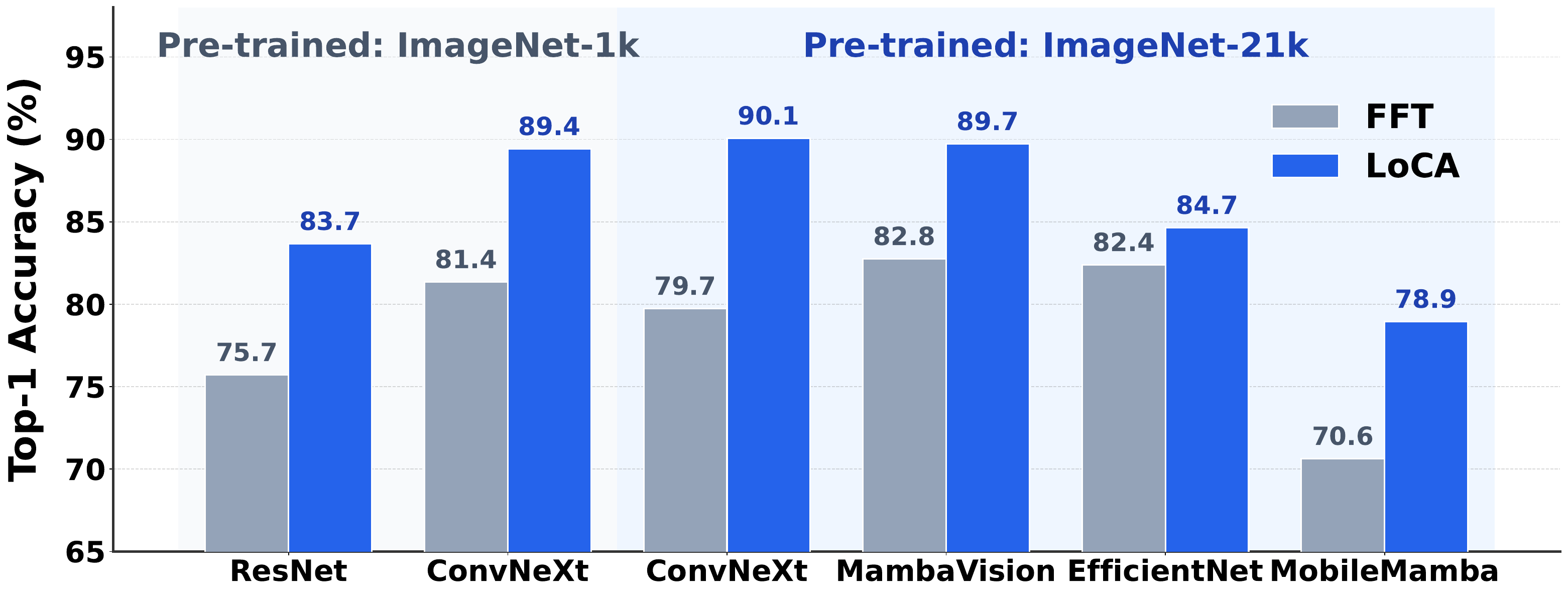}
    \vspace{-2.0em}
    \caption{Performance comparison with FFT across various backbone architectures}
    \label{fig:backbones_fgvc}
\end{wrapfigure}
against FFT on convolution-based VFM backbones, including ResNet, ConvNeXt, MambaVision, EfficientNet, and MobileMamba. 
All experiments are conducted using base models. \cref{fig:backbones_fgvc} shows that LoCA outperforms FFT across diverse backbones. 
On MobileMamba, LoCA improves Top-1 accuracy over FFT, increasing it from 70.6\% to 78.9\%.
The results demonstrate consistent gains across architectures that incorporate convolution operations.
Detailed results are provided in Appendix B.

\subsection{Generative Generalization with DreamBooth}
\label{sec:gen}

\noindent\textbf{Experimental Setup.}
For the generative task, comparative experiments involving DreamBooth \cite{dreambooth}, LoRA \cite{lora}, and FSF \cite{chen2025large} all utilize Stable Diffusion v1.4 \cite{hrnet} under identical training configurations. 
Performance analysis follows the DreamBooth evaluation protocol by using images generated from 25 prompts. 
Quantitative assessment focuses on text alignment, where DINO \cite{caron2021emerging} and CLIP-I \cite{radford2021learning} measure subject fidelity while CLIP-T \cite{radford2021learning} evaluates text prompt fidelity. 
CLIP-I and DINO calculate the average cosine similarity between the embeddings of generated and ground truth images based on CLIP and ViT-S/16-based DINO, respectively. 
Similarly, CLIP-T computes the average cosine similarity between the text prompt and the generated image embeddings.

\begin{figure}[t]
    \centering
    \setlength{\fboxsep}{0pt}
    \includegraphics[width=\linewidth]{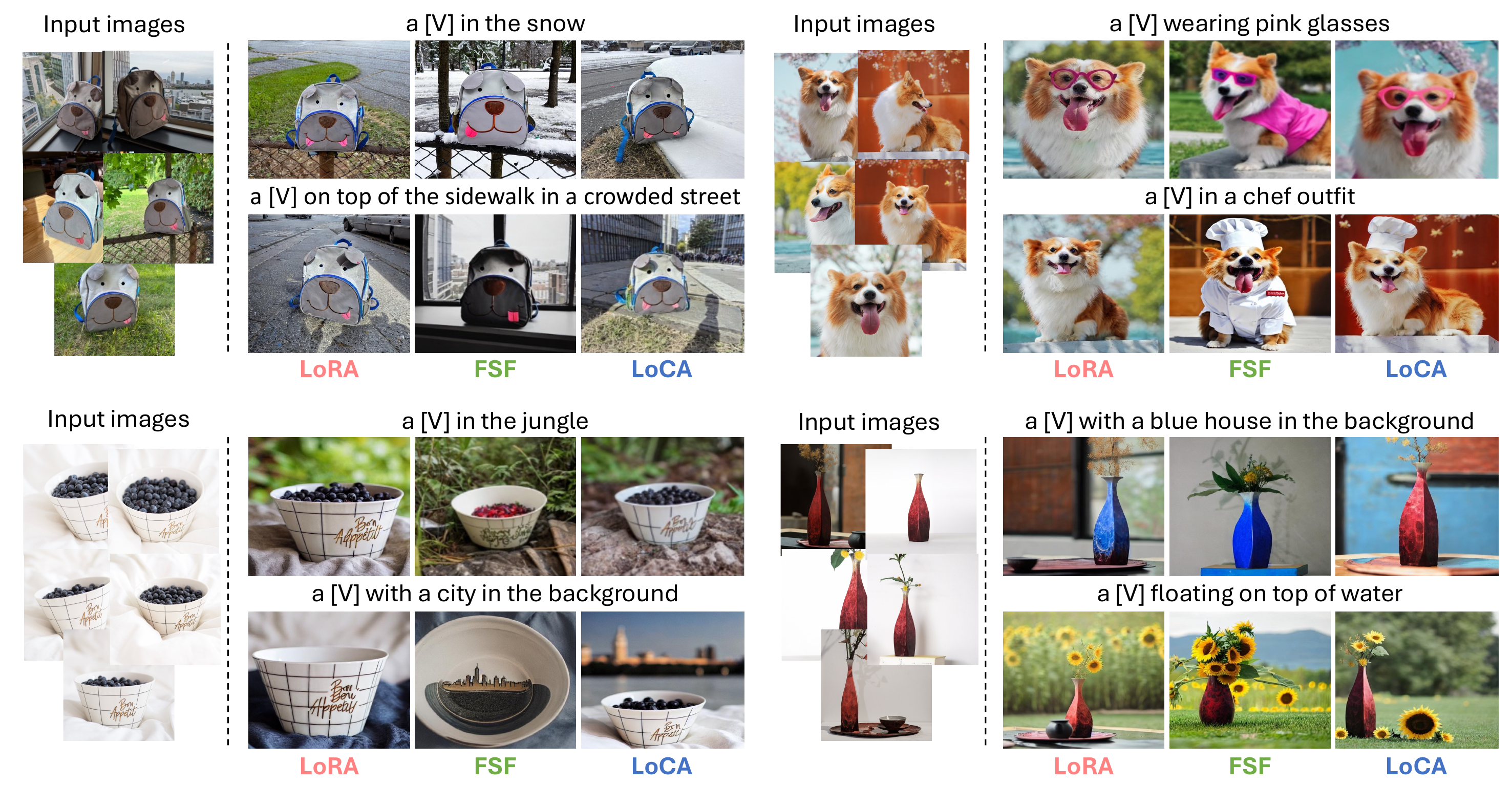}
    \caption{Qualitative results of subject-driven task. We visualize the results to compare PEFT methods: \colorbox{lorared}{\strut LoRA} \cite{lora}, \colorbox{sfgreen}{\strut FSF} \cite{chen2025large}, and \colorbox{oursblue}{\strut LoCA}.}
    \label{fig:dreambooth_visual}
    \vspace{-10pt}
\end{figure}

\noindent\textbf{Results.}
We evaluate LoCA against PEFT approaches on subject-driven generation.
\cref{tab:dreambooth} shows that LoCA achieves strong subject alignment and competitive text alignment.
LoCA outperforms LoRA by achieving the best DINO score and tying for the best CLIP-I score, while maintaining a competitive CLIP-T score.
Convolution-based adaptation incorporates spatial inductive biases for subject alignment, whereas LoRA applies low-rank updates to linear projections.

\noindent\textbf{Visualization.}
\cref{fig:dreambooth_visual} shows that LoCA better retains the structural identity of the reference subject.
For qualitative evaluation, we visualize results from LoRA \cite{lora}, FSF \cite{chen2025large}, and LoCA across four classes. 
LoRA preserves subject identity but demonstrates limited reflection of textual attributes (e.g., `chef outfit' or `city in the background').
FSF \cite{chen2025large} captures textual attributes but struggles to preserve subject identity and capture fine-grained visual details. 
For the vase-class prompt `a [V] floating on top of water', FSF generates a bulky bottle-like object rather than the intended subject.
In contrast, LoCA preserves subject identity while better reflecting textual attributes.
Detailed qualitative comparisons appear in Appendix C.

\begin{table}[t]
    \centering
    \caption{Performance comparison on synthetic-to-real DGSS using various backbones and model sizes. Models are trained on GTAV and evaluated on Cityscapes, BDD100K, and Mapillary.}
    \vspace{-1em}
    \resizebox{0.9\linewidth}{!}{
\begin{tabular}{c c c c c|c c c c}
\toprule
Method & Backbone & Param. & Trainable Param. & GFLOPs & Citys. & BDD & Map. & Avg. \\
\midrule

\multicolumn{9}{c}{\textbf{DINO Pre-trained}} \\
\addlinespace[3pt]

\rowcolor{fftgray} FFT  & ViT-B & 86.5M & 86.5M & 216 & 60.84 & 52.98 & 62.12 & 58.65 \\

SoMA & ViT-B & 86.5M & 2.3M & 216 & 66.71 & 57.48 & 67.34 & 63.84 \\
\midrule
\rowcolor{fftgray} FFT & ConvNeXt-B & 87.56M  & 87.56M & 81 & 62.18 & 57.01 & 65.00 & 61.40 \\
LoRA (Linear) & ConvNeXt-B & 87.56M & 2.9M & 81 & 63.90 & 57.87 & 65.53 & 62.43 \\
LoRA & ConvNeXt-B & 87.56M & 17.2M & 81 & 64.17 & 56.98 & 65.74 & 62.30 \\
SoMA & ConvNeXt-B & 87.56M & 2.9M & 81 & 64.99 & 57.65 & 65.67 & 62.77 \\
CA & ConvNeXt-B & 87.56M & 2.3M & 81 & 59.94 & 56.47 & 63.48 & 59.96 \\
CoLoRA & ConvNeXt-B & 87.56M & 2.2M & 81 & 61.87 & 55.70 & 64.04 & 60.53 \\
FSF & ConvNeXt-B & 87.56M & 0.6M & 81 & 60.15 & 56.98 & 63.70 & 60.27 \\
\rowcolor{oursblue} LoCA & ConvNeXt-B & 87.56M & 3.4M & 81 & \textbf{66.46} & \textbf{58.53} & 66.29 & \textbf{63.76} \\
LoCA$^{\ddagger}$ & ConvNeXt-B & 87.56M & 3.4M & 81 & 65.56 & 58.02 & \textbf{66.39} & 63.32 \\
\midrule

\rowcolor{fftgray} FFT  & ConvNeXt-L & 196.2M & 196.2M & 152 & 65.50 & 59.10 & 67.01 & 63.87 \\
LoRA (Linear) & ConvNeXt-L & 196.2M & 4.3M & 152 & 66.95 & 60.46 & 68.45 & 65.29 \\
LoRA & ConvNeXt-L & 196.2M & 25.9M & 152 & 65.74 & 60.50 & 68.62 & 64.95 \\
SoMA & ConvNeXt-L & 196.2M & 4.3M & 152 & 68.85 & 60.27 & 69.26 & 66.13 \\
CA & ConvNeXt-L & 196.2M & 4.5M & 152 & 63.16 & 58.41 & 67.32 & 62.96 \\
CoLoRA & ConvNeXt-L & 196.2M & 3.6M & 152 & 64.51 & 59.56 & 67.20 & 63.59 \\
FSF & ConvNeXt-L & 196.2M & 0.8M & 152 & 66.57 & 58.52 & 66.96 & 64.02 \\
\rowcolor{oursblue} LoCA & ConvNeXt-L & 196.2M & 5.0M & 152 & 68.54 & 61.60 & 69.33 & 66.49 \\
LoCA$^{\ddagger}$ & ConvNeXt-L & 196.2M & 5.0M & 152 & \textbf{69.73} & \textbf{62.03} & \textbf{70.62} & \textbf{67.46} \\ \midrule

\multicolumn{9}{c}{\textbf{ImageNet21k Pre-trained}} \\
\addlinespace[3pt]

FFT  & ResNet101 & 42.3M & 42.3M & 42 & 41.29 & 44.29 & 48.79 & 44.79 \\
SoMA$^\dagger$ & ResNet101 & 42.3M & 2.5M & 42 & 41.23 & 45.57 & \textbf{49.71} & 45.50 \\
LoCA & ResNet101 & 42.3M & 2.8M & 42 & \textbf{44.82} & \textbf{46.13} & 49.21 & \textbf{46.72} \\
\midrule
FFT  & MambaVision-B & 96.7M & 96.7M & 211 & 36.05 & 30.13 & 31.39 & 32.52 \\
LoCA & MambaVision-B & 96.7M & 2.5M & 211 & \textbf{45.21} & \textbf{41.68} & \textbf{45.70} & \textbf{44.20} \\
\midrule
FFT  & MambaVision-L3 & 737.5M & 737.5M & 1,556 & 52.88 & 45.87 & 56.07 & 51.61 \\
LoCA & MambaVision-L3 & 737.5M & 9.0M & 1,556 & \textbf{59.94} & \textbf{50.61} & \textbf{61.39} & \textbf{57.31} \\


\bottomrule
\label{tab:gtav}
\end{tabular}}
    \parbox{\linewidth}{\footnotesize
    $^{\dagger}$ PEFT via linearization of patch-level convolutions and weights.\\
    $^{\ddagger}$ Channel mixing path initialization follows the SVD-based approach of SoMA \cite{yun2025soma}.}
    \label{tab:dgg}
\end{table}

\subsection{Domain Generalization for Semantic Segmentation}
\label{sec:dg}

\noindent\textbf{Experimental Setup.}
Experiments utilize convolution-based backbones, specifically ResNet \cite{resnet}, and ConvNeXt \cite{woo2023convnext}. 
Evaluation further includes recent vision foundation models with hybrid transformer–convolution architectures, including DINOv3-ConvNeXt \cite{dinov3} and MambaVision \cite{mambavision}. 
All models employ a consistent Mask2Former \cite{mask2former} decoder for DGSS and DGOD tasks. 
A fixed rank of 16 across all experiments ensures a fair comparison. 
For DGSS and DGOD tasks, models are trained on the GTAV dataset and evaluated on three real-world benchmarks: Cityscapes \cite{cityscapes}, BDD100K \cite{bdd100k}, and Mapillary \cite{mapillary}. 
Appendix D presents detailed experimental settings and performance analysis for different backbones.

\begin{figure}[t]
    \centering
    \includegraphics[width=0.95\linewidth]{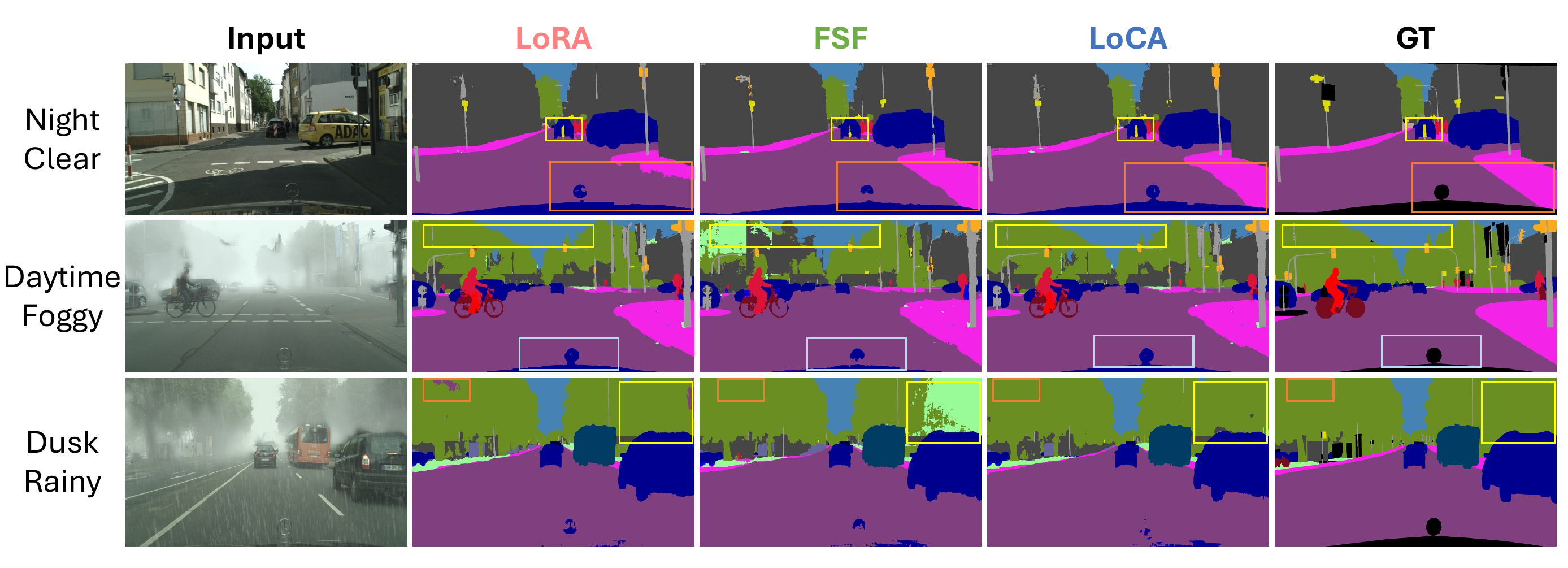}
    \vspace{-1em}
    \caption{Qualitative domain adaptation results (GTAV → Cityscapes)}
    \label{fig:dgg}
\end{figure}

\noindent\textbf{Results.}
As shown in~\cref{tab:gtav}, we evaluate the generalization capability of LoCA on DGSS by training on GTAV and testing on real-world out-of-distribution (OOD) benchmarks.
LoCA with DINO-pretrained ConvNeXt backbones \cite{dinov3} demonstrates competitive performance compared to other PEFT methods on OOD datasets. 
LoRA-based adaptation flattens convolution kernels and requires learning a large number of parameters. 
LoCA achieves strong performance with substantially fewer trainable parameters than LoRA. 
Initializing low-rank components with the principal singular components following SoMA boosts performance by approximately $+1.0$ on average across three datasets in ConvNeXt-L. Conversely, zero initialization outperforms this method for ConvNeXt-B. The gain observed in ConvNeXt-L suggests that larger models retain more salient eigenvalue structures from pre-trained weights.
Benchmarking against SoMA results on ViT-B \cite{vit}, we evaluated ConvNeXt-B with a comparable parameter count. ConvNeXt-B achieves performance parity with ViT-B at 81 GFLOPs, representing a $2.6 \times$ reduction from the 216 GFLOPs required by ViT-B. The ConvNeXt-B-based LoCA maintains high accuracy while mitigating computational overhead.
For DGSS, models are trained on GTAV and evaluated on Cityscapes, BDD100K, and Mapillary. Detailed DGOD results are provided in Appendix D.

\noindent\textbf{Visualization.} 
Qualitative evaluations utilize ConvNeXt-B to compare LoRA \cite{lora}, FSF \cite{chen2025large}, and LoCA on night clear, foggy, and rainy Cityscapes scenes \cite{cityscapes}. 
These challenging environments serve as a benchmark for robustness. 
As shown in \cref{fig:dgg}, our method generates fine-grained segmentation maps in both rainy and foggy scenes. 
LoCA preserves sharp object boundaries and captures fine-grained details in the boxed regions. 
LoCA exhibits stronger robustness to weather-induced noise than LoRA and FSF.

\noindent\textbf{Generalization across Backbones.}
Recent vision foundation models adopt convolutional architectures such as Mamba-based designs \cite{gu2024mamba}. We adapt convolutional components during fine-tuning of MambaVision \cite{mambavision}, ResNet \cite{resnet}, and ConvNeXt \cite{convnext}. 
On MambaVision-B, LoCA improves the average score from 32.52 to 44.20, corresponding to $+11.68$ points or a 35.9\% relative gain over FFT.
\cref{tab:dgg} further shows that LoCA achieves competitive performance while using only about 2.5\% of the parameters.

\subsection{Ablation Studies}
\label{sec:ablation}

\begin{figure}[t]
    \centering
    \includegraphics[
        width=\linewidth,
        trim=0 0 0 0,
        clip
    ]{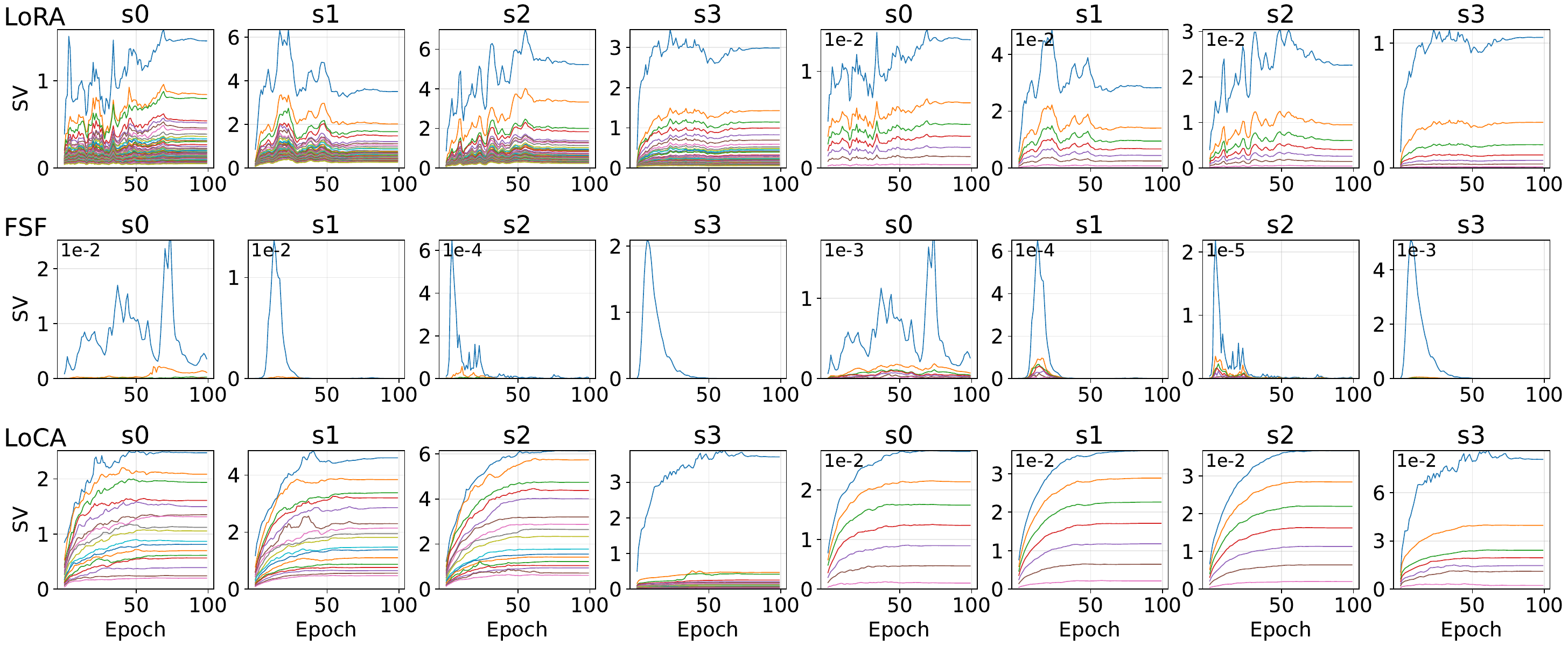}
    \vspace{-2em}
    \caption{
    Singular value evolution of ConvNeXt-B depthwise convolution weights across three PEFT methods: LoRA, FSF, and LoCA. The y-axis denotes singular values. Columns 1--4: channel-wise SVD of depthwise weights; columns 5--8: spatial SVD. Stable distributions prevent spectral tail growth and ensure controlled effective rank behavior.}
    \label{fig:sv}
\end{figure}

\noindent\textbf{Evolution of Representational Capacity.}
The evolution of singular value spectra during training reflects representational capacity during adaptation. 
A distributed singular value spectrum can indicate effective representation learning, with information distributed across multiple components rather than concentrated in a few dominant ones \cite{yunis2024rank}. 
To analyze this spectral evolution, we track the singular value spectra of a ConvNeXt-B depthwise convolution weight matrix (dwconv) on VTAB-1k Oxford Pets. 
As shown in \cref{fig:sv}, LoRA and FSF exhibit limited variation in their basis components across training epochs. 
In contrast, LoCA exhibits progressively diverse basis components with smooth singular value growth across both leading and trailing components. This spectral evolution reflects kernel importance identified through covariance analysis and enables joint channel–spatial adaptation.

\begin{figure}[t]
    \centering
    \includegraphics[width=0.8\linewidth]{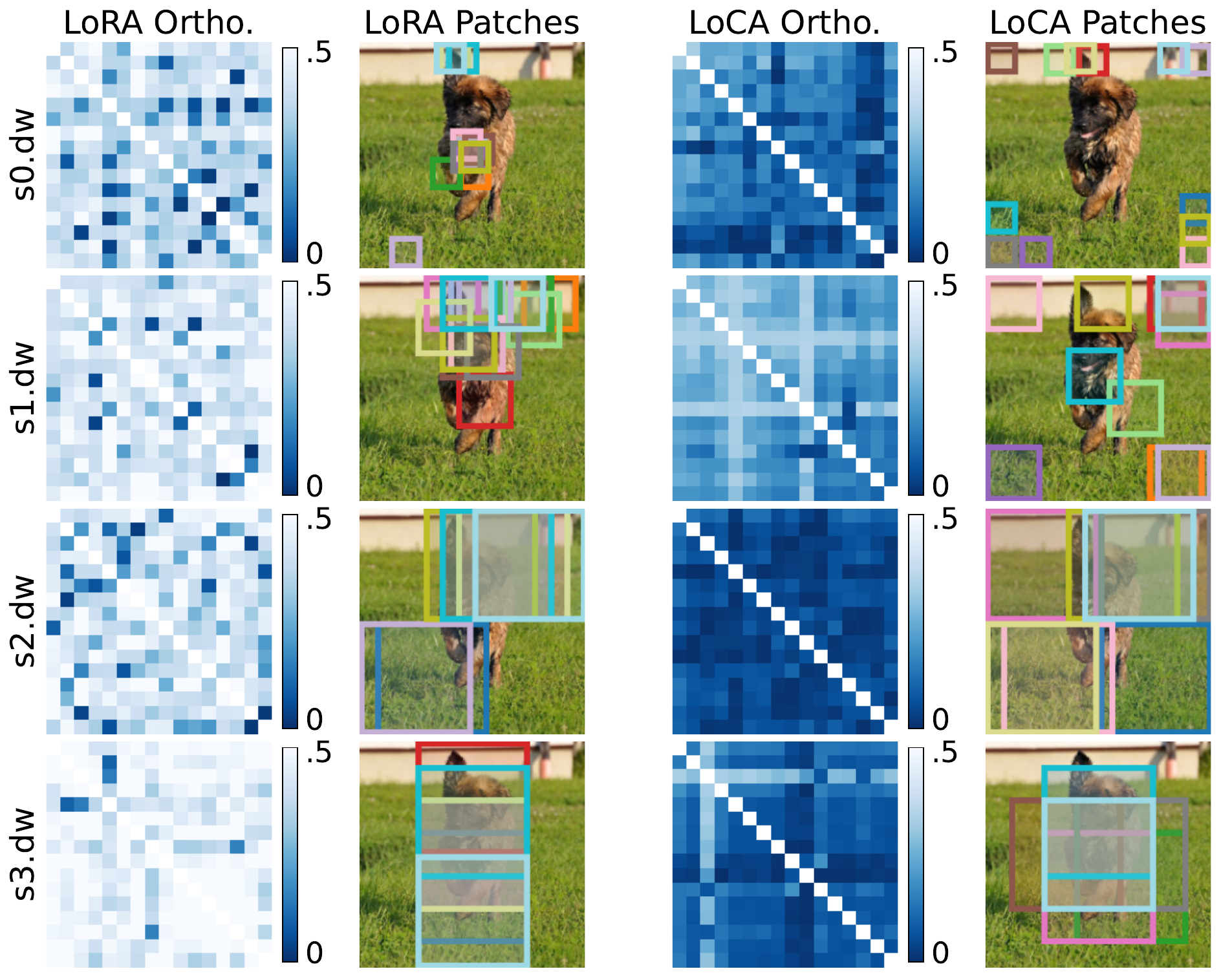}
    \vspace{-4mm}
    \caption{Absolute pairwise cosine similarity and top-activated localized patches \cite{Visualizing} of trained rank components across stages (s0--3). Localized patches denote image regions with the strongest activation responses for each rank component.}
    \vspace{-4mm}
    \label{fig:expressivity}
\end{figure}

\noindent\textbf{Spatial Representation Analysis.}
We validate the effectiveness of LoCA in convolutional adaptation by visualizing the orthogonality and spatial coverage of rank components. \cref{fig:expressivity} shows the cosine similarity among rank components and their spatial coverage. LoRA rank components exhibit higher similarity and tend to collapse into redundant feature directions. In contrast, LoCA produces more orthogonal rank components, which encourages diverse subspace representations and improves convolutional expressivity. The increased diversity of rank components becomes more evident at the patch level. The highly activated patches in LoCA are distributed across diverse spatial regions rather than localized to a single spatial cue. LoRA focuses on localized regions, whereas LoCA learns orthogonal rank components with broader spatial coverage.

\noindent\textbf{Ablation Analysis of Proposed Methods.}
We analyze the contribution of each proposed method relative to a LoRA baseline. The analysis compares
\begin{wraptable}{r}{0.6\columnwidth}
\centering
\vspace{-2.4em}
\vspace{-1em}
\caption{Ablation study of proposed methods under DGSS benchmarks}
\label{tab:ablation}
\resizebox{\linewidth}{!}{
\begin{tabular}{l|cccc}
\toprule
Tuning Method 
& Citys. 
& BDD 
& Map. 
& Avg. \\
\midrule
LoRA Linear
& 66.95 & 60.46 & 68.45 & 65.29 \\

$\llcorner$ LoRA Convolution
& 65.74 {\scriptsize \textcolor{blue}{$\triangledown$1.21}} 
& 60.50 {\scriptsize \textcolor{red}{$\triangle$0.04}} 
& 68.62 {\scriptsize \textcolor{red}{$\triangle$0.17}} 
& 64.95 {\scriptsize \textcolor{blue}{$\triangledown$0.34}} \\
\midrule

$\llcorner$ Channel Mixing
& 68.22 {\scriptsize \textcolor{red}{$\triangle$1.27}} 
& 61.12 {\scriptsize \textcolor{red}{$\triangle$0.66}} 
& 69.13 {\scriptsize \textcolor{red}{$\triangle$0.68}} 
& 66.16 {\scriptsize \textcolor{red}{$\triangle$0.87}} \\

$\llcorner$ Spatial Basis 
& 69.44 {\scriptsize \textcolor{red}{$\triangle$2.49}} 
& 61.39 {\scriptsize \textcolor{red}{$\triangle$0.93}} 
& 70.26 {\scriptsize \textcolor{red}{$\triangle$1.81}} 
& 67.03 {\scriptsize \textcolor{red}{$\triangle$1.74}} \\

$\llcorner$ Hierarchical Rank
& 69.73 {\scriptsize \textcolor{red}{$\triangle$2.78}} 
& 62.03 {\scriptsize \textcolor{red}{$\triangle$1.57}} 
& 70.62 {\scriptsize \textcolor{red}{$\triangle$2.17}} 
& 67.46 {\scriptsize \textcolor{red}{$\triangle$2.17}} \\
\bottomrule
\end{tabular}}
\vspace{-2.3em}
\end{wraptable}
incremental variants, including standard convolutional LoRA, channel mixing, spatial basis refinement, and hierarchical rank scheduling.
As shown in \cref{tab:ablation}, applying the standard convolutional LoRA formulation decreases the average score by $0.34$ points.
Naively applying LoRA to convolutional kernels does not improve adaptation performance and may degrade OOD performance.
In contrast, the proposed channel mixing increases the average score to $66.16$, yielding a $+0.87$ gain over LoRA Linear.
Adding spatial basis refinement further improves the average score to $67.03$, yielding a $+1.74$ gain over LoRA Linear.
Hierarchical rank scheduling achieves the best average score of $67.46$, corresponding to a $+2.17$ gain over LoRA Linear.
Convolutional architectures encode coarse-to-fine information across stages.
Hierarchical rank scheduling aligns the adaptation capacity with this structural property.

\noindent \textbf{Analysis of Covariance-SVD Initialization.} We compare covariance-SVD initialization with other initialization strategies on VTAB-1k and DGSS data. 
Across diverse tasks, covariance-SVD initialization consistently outperforms other 
\begin{wraptable}{r}{0.6\textwidth}
\vspace{-12.2mm}
\centering
\caption{Ablation study on initialization strategies}
\label{tab:init_ablation}
\resizebox{\linewidth}{!}{
\begin{tabular}{l|cccc}
\hline
Initialization & Zero & Flatten SVD & Uniform & Covariance \\
\hline
VTAB-1k Avg. & 75.7 & 73.0 & 75.7 & \textbf{75.9} \\
DGSS Avg.    & 65.6 & 66.2 & 66.3 & \textbf{66.5} \\
\hline
\end{tabular}
}
\vspace{-3em}
\end{wraptable}
initialization strategies. The consistent gains suggest that preserving the directional structure of convolutional feature subspaces benefits representation learning. Covariance-SVD preserves pre-trained spatial priors because SVD is applied to spatial covariance rather than a flattened convolution tensor.
Covariance-SVD performs best on both VTAB-1k and DGSS (\cref{tab:init_ablation}).

\section{Conclusions}
Although LoRA is the dominant PEFT approach, convolutional adaptation remains underexplored despite the centrality of spatial information to visual adaptation.
To this end, we present Low-Rank Convolutional Adaptation (LoCA), a convolution-aware PEFT framework for adapting vision foundation models.
LoCA decouples the adaptation into low-rank channel adaptation and spatial basis refinement.
This convolution-aware PEFT framework addresses spatial-channel entanglement by decoupling channel and spatial adaptation while preserving pre-trained spatial priors.
Furthermore, our hierarchical rank scheduling aligns adaptation capacity with the backbone’s hierarchical feature extraction.
Experimental results show that LoCA achieves strong performance across tasks and backbones, outperforming existing methods on several benchmarks and remaining competitive on others.

\section*{Acknowledgements}
This research was supported by the Institute of Information \& Communications Technology Planning \& Evaluation (IITP) grant funded by the Korea government (MSIT) (No. RS-2019-II190079, Artificial Intelligence Graduate School Program (Korea University), 1\%; No. RS-2025-25439490, 40\%), Culture, Sports and Tourism R\&D Program through the Korea Creative Content Agency grant funded by the Ministry of Culture, Sports and Tourism in 2024 (No. RS-2024-00345025, International Collaborative Research and Global Talent Development for the Development of Copyright Management and Protection Technologies for Generative AI, 10\%), the National Research Foundation of Korea (NRF) grant funded by the Korea government (MSIT) (No. RS-2024-00341514, 39\%), the Industrial Technology Innovation Program (No. RS-2025-25448266, Development of Humanoid Robots Specialized in Display Manufacturing Processes Based on AI Foundation Models, 10\%) grant funded by the Korea government (MOTIE).

\clearpage  

\bibliographystyle{splncs04}
\bibliography{main}

 



\title{Supplementary Material\\LoCA: Spatially-Aware Low-Rank Convolutional Adaptation of Vision Foundation Models} 

\titlerunning{LoCA}

\author{
Sojung An\inst{1}\equalcontrib,\,
Junha Lee\inst{2,3}\equalcontrib,\,
Sujeong You\inst{2},\,
Nam Ik Cho\inst{3},\,
Donghyun Kim\inst{1}\corrauthor
}

\authorrunning{S.~An et al.}
\institute{
Korea University, Seoul, Republic of Korea
\and
Korea Institute of Industrial Technology, Ansan, Republic of Korea
\and
Seoul National University, Seoul, Republic of Korea\\[0.5em]
$^{*}$Equal contribution
$^{\dagger}$Corresponding author: \email{d\_kim@korea.ac.kr}\\
\raisebox{-\mydepth}{\includegraphics[height=1.\myheight]{figs/github.png}}
\textbf{\url{https://github.com/ssojungan/loca}}
}

\maketitle

\appendix
\renewcommand{\thefigure}{\Alph{figure}}
\renewcommand{\thetable}{\Alph{table}}
\setcounter{figure}{0}
\setcounter{table}{0}

\noindent This supplementary material is organized as follows:
\begin{itemize}
    \item \cref{sec:cost} provides the computational cost of LoRA and LoCA for different convolutional operator types and input resolutions.
    \item \cref{sec:fgvc} presents comprehensive VTAB-1k and FGVC results, including training settings omitted from Sec.~5.1.
    \item \cref{sec:generative} specifies the DreamBooth training protocol underlying the generative experiments in Sec.~5.2.
    \item \cref{sec:dg2} describes the settings and benchmark configurations for the domain generalization experiments in Sec.~5.3.
    \item \cref{sec:additional} offers additional analyses of representational capacity through singular value evolution.
    \item \cref{sec:additional_results} provides additional experiments for analyzing the sensitivity of LoCA.
    \item \cref{sec:code} clarifies the formulations in Sec.~4 through simplified code implementations of LoRA and LoCA.
\end{itemize}

\section{Computational Cost}
\label{sec:cost}
This section analyzes the computational cost of LoRA and LoCA.
Parameter-Efficient Fine-Tuning (PEFT) aims to achieve strong performance with a small number of adaptation parameters. We therefore report the parameter complexity and FLOPs.

\noindent\textbf{Computational Cost Formulation w.r.t. Kernel Size.}
The parameter complexity ($P$) of LoRA is approximated as
$P_{\text{LoRA}} \approx r \cdot k^{2} \cdot C$,
where $r$, $k$, and $C$ denote the adaptation rank, the kernel size, and the channel dimension, respectively.
LoCA decomposes the adaptation into channel mixing and spatial basis components,
$P_{\text{LoCA}} \approx r \cdot C + \text{const}(k^{4} + k^{2})$.
The first term represents channel mixing parameters, and the remaining terms correspond to spatial basis parameters. 
The channel-dependent term scales with $k^{2}C$ in LoRA and $C$ in LoCA.
Implementation details are provided in Sec.~\ref{sec:code}.

\begin{figure}
    \centering
    \begin{tikzpicture}
        \node[anchor=south west,inner sep=0] (img) at (0,0)
        {\includegraphics[width=\linewidth]{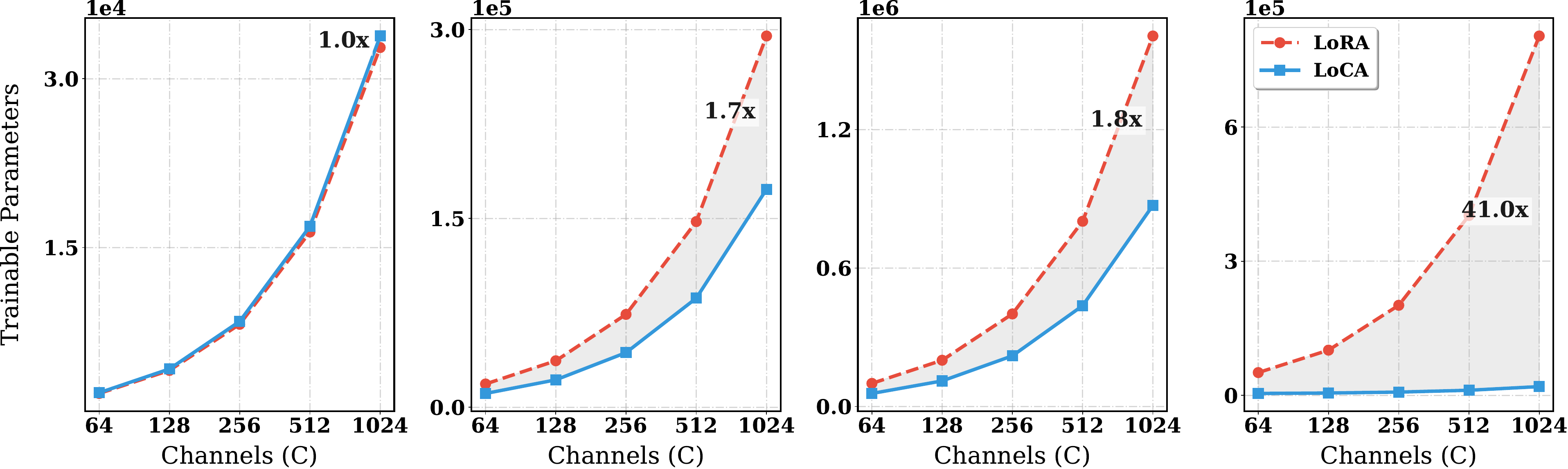}};
        \begin{scope}[x={(img.south east)},y={(img.north west)}]
            \node at (0.15,-0.1) {\scriptsize (a) K=1};
            \node at (0.38,-0.1) {\scriptsize (b) K=3};
            \node at (0.64,-0.1) {\scriptsize (c) K=7};
            \node at (0.89,-0.1) {\scriptsize (d) K=7 (Depthwise)};
        \end{scope}
        \end{tikzpicture}
    \vspace{-1em}
    \caption{Computational cost with respect to kernel size and channel dimension}
    \label{fig:peft_comparison}
\end{figure}
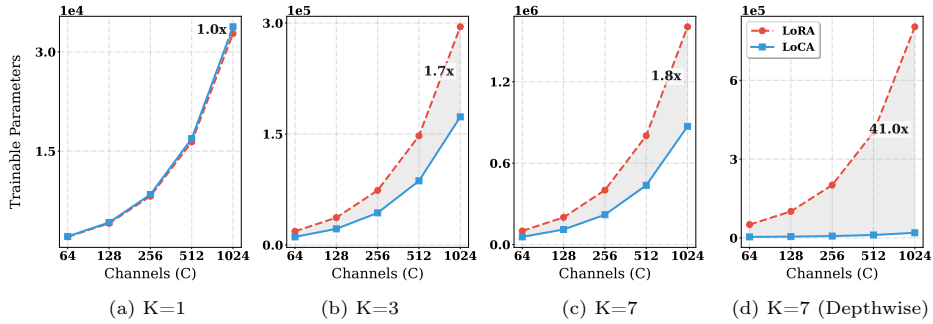

\noindent\textbf{Parameter Complexity Analysis.}
\cref{fig:peft_comparison} illustrates parameter complexity relative to kernel size. For $1\times1$ kernels, LoCA incurs a marginal parameter overhead ($\approx 1.0\times$) compared to LoRA due to the spatial basis. However, increasing kernel sizes amplifies the parameter efficiency of LoCA. LoRA exhibits quadratic complexity growth ($k^{2}$) with the kernel size due to its reliance on flattened-kernel decomposition. In contrast, LoCA decouples channel mixing and spatial basis components, isolating channel-dependent parameters from the kernel area.
This efficiency gain is particularly evident in the depthwise convolution of ConvNeXt (\cref{fig:peft_comparison}d). In this configuration, LoCA constructs a shared spatial basis $s_{basis}$ defined solely by the kernel size $k$, with individual channels utilizing specific coefficients. While LoRA's complexity involves a multiplicative $r \cdot k^{2} \cdot C$ factor, LoCA confines $k$-dependent terms to the spatial basis ($k^{4}$ and $k^{2}$), maintaining a channel-dependent term of only $r \cdot C$. Sharing spatial parameters across channels ensures superior parameter scaling in large-kernel operators, such as the $7\times7$ depthwise convolution.

\noindent\textbf{FLOPs.} Both LoRA and LoCA utilize an identical convolution operator during inference. Therefore, inference FLOPs remain equivalent, with the distinction residing solely in parameter complexity.

\section{Fine-Grained Classification}
\label{sec:fgvc}

We provide an extensive experimental analysis in Sec. 5.1, reporting task-wise results on VTAB-1k tasks and benchmarks on four FGVC datasets.
VTAB-1k contains 19 tasks that cover a broad spectrum of domains and semantics.
These are grouped into three sets: \taskNatural{} (Caltech101, CIFAR-100, DTD, Flowers102, Pets, Sun397, SVHN), \taskSpecialized{} (EuroSAT, Resisc45, Patch Camelyon, Retinopathy), and \taskStructured{} (Clevr/count, Clevr/distance, dSprites/location, dSprites/orientation, SmallNORB/azimuth, SmallNORB/elevation, DMLab, KITTI/distance).
The FGVC benchmark contains four specialized datasets aimed at fine-grained classification: CUB200, Stanford Dogs, Stanford Cars, and NABird.

\subsection{Setup Details}
For a fair comparison, we follow the preprocessing protocol of CA~\cite{convadapter} and use Center Crop, following~\cite{centrecrop}. 
For FGVC, we additionally apply RandomResizedCrop with a minimum scale of 0.2 and Horizontal Flip~\cite{horizontalflip}. 
For few-shot classification, we use the same augmentation policy as in FGVC and adopt the hyperparameter search range listed in \cref{tab:fg_hyper}.

\begin{table}[!h]
\centering
\caption{Hyperparameters for image classification tasks of FGVC and VTAB-1k}
\label{tab:fg_hyper}
\resizebox{0.3\columnwidth}{!}{%
\begin{tabular}{@{}ll@{}}
\toprule
             & \multicolumn{1}{c}{All Backbones}      \\ \midrule
Optimizer    & AdamW \cite{adamw} \\
Learning rate     & 1e-3 \\
Weight decay     & 1e-4  \\
LR schedule  & cosine \\
Total Epochs & 100    \\
Warmup       & 10     \\ \bottomrule
\end{tabular}%
}
\end{table}

\subsection{Experimental Results}

In \cref{tab:vtab_convnext_details} and \cref{tab:vtab_resnet_details}, our task-wise analysis shows that LoCA consistently performs well across VTAB-1k tasks.
We highlight the top-1, top-2, and top-3 entries in each column with progressively darker shades. 
For the \taskNatural{} group, where the target distribution is closest to ImageNet pre-training, LoCA is competitive with other PEFT baselines while achieving top results on several tasks. 
This indicates that LoCA can improve performance on \taskNatural{} tasks where PEFT baselines tend to lag (e.g., DTD/SVHN), without sacrificing overall competitiveness under limited domain shift.
Notably, LoCA achieves the highest performance across all tasks within the \taskSpecialized{} group. 
Because this group possesses domain-specific characteristics distinct from those of natural images, these results demonstrate that our approach effectively adapts to fine-grained, specialized domains. 
For the \taskStructured{} group, LoCA outperforms LoRA. 
We observe high performance variance in the \taskStructured{} group when the official LoRA implementation jointly tunes convolutional and linear layers. 
This variance likely occurs because these tasks rely heavily on encoding spatial and shape information. 
While some classes show gains over the linear-only configuration, others experience significant performance degradation.

In \cref{tab:fgvc_convnext_details} and \cref{tab:fgvc_resnet_details}, LoCA achieves the best mean accuracy on both backbones while remaining parameter-efficient. 
Using ConvNeXt-B pre-trained on ImageNet-21K, LoCA reaches the highest average accuracy, outperforming LoRA with significantly fewer trainable parameters (3.7 M vs.\ 17.6 M). 
The performance gains on NABird and CUB200 demonstrate our method's ability to capture subtle inter-class differences in fine-grained tasks. 
FSF performs best on Stanford Dogs. 
LoCA achieves the best results on CUB200 and NABird, matches the top accuracy on Stanford Cars, and obtains the best overall mean with 3.7M trainable parameters.
Under the ResNet-50 architecture pre-trained on ImageNet-1K, LoCA yields the highest mean accuracy with only 1.9 M trainable parameters, surpassing both LoRA and FSF. 
Notably, LoCA improves CUB200 and NABird, indicating robust adaptation even with a smaller backbone.
We also observe that full fine-tuning underperforms PEFT methods under the low-data regime, consistent with the tendency to overfit when updating all parameters.

\begin{table}[h!]
\caption{Performance comparisons on the VTAB-1k benchmark with ConvNeXt-B models pre-trained on ImageNet-21K}
\vspace{-1em}
\centering
\resizebox{0.99\columnwidth}{!}{
\begin{tabular}{c|ccccccc|cccc|cccccccc|cc}
\toprule
& \multicolumn{7}{c|}{Natural}
& \multicolumn{4}{c|}{Specialized}
& \multicolumn{8}{c|}{Structured}
& & \\ \midrule

& \rotatebox{90}{Caltech101}
& \rotatebox{90}{CIFAR100}
& \rotatebox{90}{DTD}
& \rotatebox{90}{Flowers102}
& \rotatebox{90}{Pets}
& \rotatebox{90}{SVHN}
& \rotatebox{90}{Sun397}

& \rotatebox{90}{Patch Camelyon}
& \rotatebox{90}{EuroSAT}
& \rotatebox{90}{Resisc45}
& \rotatebox{90}{Retinopathy}

& \rotatebox{90}{Clevr/count}
& \rotatebox{90}{Clevr/distance}
& \rotatebox{90}{DMLab}
& \rotatebox{90}{KITTI}
& \rotatebox{90}{dSprites/loc}
& \rotatebox{90}{dSprites/ori}
& \rotatebox{90}{SmallNORB/azi}
& \rotatebox{90}{SmallNORB/ele}

& \rotatebox{90}{Mean}
& \rotatebox{90}{Params. (M)} \\ \midrule

Full fine-tuning
& \second{91.0} & \third{66.2} & 74.8 & \second{99.6} & 92.0 & \second{88.5} & 51.5
& \first{86.8} & \second{95.9} & \second{88.7} & \first{78.8}
& 81.6 & 53.6 & \first{55.3} & \second{82.4}
& \second{95.0} & \second{70.3} & \third{37.2} & \third{35.4}
& 73.7 & 87.7 \\


LoRA
& \third{91.0} & 66.0 & \third{75.2} & \third{99.6} & 92.2 & \first{89.5} & 52.7
& 85.7 & 95.4 & \third{87.8} & \third{78.4}
& 81.0 & 49.4 & \second{55.3} & 79.5
& \first{96.9} & \third{69.4} & \second{37.4} & 30.7
& \third{74.4} & 17.3 \\

CA
& 90.9 & 66.0 & 74.9 & 98.8 & \third{92.4} & 52.9 & \first{88.4}
& \third{86.0} & \third{95.6} & 85.7 & 77.9
& \second{86.5} & \third{59.5} & \third{55.0} & \first{93.7}
& 67.1 & \first{83.5} & \first{39.0} & 34.7
& \second{75.2} & 6.8 \\

CoLoRA
& 89.9 & 58.2 & 71.7 & 98.9 & 91.5 & 83.3 & 39.1
& 85.1 & 93.7 & 78.7 & 75.1
& \third{83.9} & \first{66.1} & 49.1 & 80.0
& 76.2 & 43.7 & 22.6 & \first{43.9}
& 70.0 & \third{4.6} \\

FSF
& \first{94.8} & \first{71.7} & \second{76.9} & 99.6 & \second{93.1} & \third{87.1} & \second{57.5}
& 85.1 & 94.6 & 87.6 & 74.8
& 70.9 & \second{62.8} & 50.3 & \third{82.7}
& 89.4 & 60.4 & 31.2 & 29.0
& 73.6 & \first{1.1} \\

LoCA
& 90.8 & \second{69.5} & \first{77.1} & \first{99.7} & \first{93.6} & 86.7 & \third{55.3}
& \second{86.6} & \first{96.1} & \first{88.8} & \second{78.5}
& \first{92.9} & 54.7 & 53.7 & \third{83.2}
& \third{89.7} & 67.4 & 37.1 & \second{41.3}
& \first{75.9} & 5.0 \\

\bottomrule
\end{tabular}}
\vspace{-8pt}
\label{tab:vtab_convnext_details}
\end{table}
\vspace{-1em}
\begin{table}[h!]
\caption{Performance comparisons on the VTAB-1k benchmark with ResNet-50 models pre-trained on ImageNet-1K}
\vspace{-1em}
\centering
\resizebox{0.99\columnwidth}{!}{
\begin{tabular}{c|ccccccc|cccc|cccccccc|cc}
\toprule
& \multicolumn{7}{c|}{Natural}
& \multicolumn{4}{c|}{Specialized}
& \multicolumn{8}{c|}{Structured}
& & \\ \midrule

& \rotatebox{90}{Caltech101}
& \rotatebox{90}{CIFAR100}
& \rotatebox{90}{DTD}
& \rotatebox{90}{Flowers102}
& \rotatebox{90}{Pets}
& \rotatebox{90}{SVHN}
& \rotatebox{90}{Sun397}

& \rotatebox{90}{Patch Camelyon}
& \rotatebox{90}{EuroSAT}
& \rotatebox{90}{Resisc45}
& \rotatebox{90}{Retinopathy}

& \rotatebox{90}{Clevr/count}
& \rotatebox{90}{Clevr/distance}
& \rotatebox{90}{DMLab}
& \rotatebox{90}{KITTI}
& \rotatebox{90}{dSprites/loc}
& \rotatebox{90}{dSprites/ori}
& \rotatebox{90}{SmallNORB/azi}
& \rotatebox{90}{SmallNORB/ele}

& \rotatebox{90}{Mean}
& \rotatebox{90}{Params. (M)} \\ \midrule

Full fine-tuning
& 83.9 & 25.9 & \third{63.4} & \second{90.4} & \second{91.0} & \second{33.1} & \third{71.4}
& 79.3 & 91.6 & \second{82.0} & 75.1
& \third{60.4} & \third{50.6} & \second{45.8} & \third{56.4}
& \second{61.9} & \second{78.6} & \third{27.3} & 37.6
& 61.0 & 23.6 \\

LoRA
& 83.88 & 26.41 & 62.61 & \third{89.15} & 90.41 & 31.82 & 71.14
& 81.55 & \third{91.65} & \third{81.32} & \third{75.62}
& \second{67.71} & 49.69 & \third{45.36} & \second{73.12}
& \first{64.56} & 76.37 & \first{28.73} & \second{44.10}
& \second{65.01} & 19.0 \\

CA
& \third{86.98} & \third{27.22} & \second{64.40} & 82.21 & 88.98 & 32.67 & 51.31
& 78.59 & 88.20 & 75.29 & 73.80
& 35.94 & 44.98 & 35.40 & 41.50
& 15.29 & 69.95 & 14.72 & 38.21
& 55.03 & \second{1.4} \\

CoLoRA
& \first{89.1} & \second{30.0} & 62.8 & 87.5 & 88.9 & 32.4 & \first{75.3}
& \third{82.6} & \second{92.2} & 81.1 & 74.3
& 54.0 & \first{54.0} & 42.1 & \first{76.4}
& 37.2 & \first{80.0} & 22.6 & \first{48.8}
& \third{63.7} & \third{1.4} \\

FSF
& \second{87.0} & 22.0 & 61.0 & 88.5 & \first{93.0} & \third{33.0} & 54.0
& \first{83.5} & 87.5 & 73.5 & \first{76.5}
& 40.5 & 48.5 & 42.5 & 12.0
& 18.0 & 67.5 & 14.5 & 26.0
& 57.0 & \first{0.7} \\

LoCA
& 84.4 & \first{30.8} & \first{65.6} & \first{90.6} & \third{91.0} & \first{34.5} & \second{73.6}
& \second{83.5} & \first{92.4} & \first{82.1} & \second{76.5}
& \first{76.6} & \second{53.2} & \first{48.2} & 50.6
& \third{59.5} & \third{77.0} & \second{27.6} & \third{43.4}
& \first{65.3} & 1.5 \\

\bottomrule
\end{tabular}}
\vspace{-8pt}
\label{tab:vtab_resnet_details}
\end{table}

\begin{table*}[t]
\centering

\begin{minipage}{0.48\textwidth}
\centering
\caption{FGVC benchmark with ConvNeXt-B (ImageNet-21K)}
\vspace{-1em}
\resizebox{0.99\linewidth}{!}{
\begin{tabular}{c|cccc|cc}
\toprule
& CUB200 & Dogs & Cars & NABird & Mean & Params. \\ \midrule
Full FT & 73.3 & 73.6 & \second{88.3} & 67.7 & 75.7 & 79.7 \\
LoRA & \second{89.0} & 86.4 & \first{92.9} & \third{84.4} & \second{88.4} & 17.6 \\
CA & 73.6 & \third{88.5} & 79.5 & 67.1 & 77.2 & 6.8 \\
CoLoRA & \third{87.5} & \third{88.5} & 83.8 & \second{84.6} & 86.1 & \third{4.6} \\
FSF & 85.5 & \first{96.1} & \third{86.3} & 84.3 & \third{88.0} & \first{1.1} \\
Ours & \first{90.5} & \second{89.1} & \first{92.9} & \first{87.8} & \first{90.1} & \second{3.7} \\
\bottomrule
\end{tabular}}
\label{tab:fgvc_convnext_details}
\end{minipage}
\hfill
\begin{minipage}{0.48\textwidth}
\centering
\caption{FGVC benchmark with ResNet-50 (ImageNet-1K)}
\vspace{-1em}
\resizebox{0.99\linewidth}{!}{
\begin{tabular}{c|cccc|cc}
\toprule
& CUB200 & Dogs & Cars & NABird & Mean & Params. \\ \midrule
Full FT & 73.3 & 73.6 & \first{88.3} & 67.7 & 75.7 & 24.1 \\
LoRA & \second{78.3} & \third{85.8} & \second{87.8} & \third{71.9} & 81.0 & \third{2.4} \\
CA & - & - & - & - & \second{83.5} & \second{1.9} \\
CoLoRA & 74.2 & 81.7 & \third{84.1} & 65.7 & 76.4 & \first{1.0} \\
FSF & \third{76.7} & \first{90.6} & 79.1 & \first{77.9} & \third{81.1} & 2.5 \\
Ours & \first{80.8} & \second{88.7} & \first{88.3} & \second{76.9} & \first{83.7} & \second{1.9} \\
\bottomrule
\end{tabular}}
\label{tab:fgvc_resnet_details}
\end{minipage}

\end{table*}


\section{Subject-Driven Text-to-Image Generation}
\label{sec:generative}
We further evaluate LoCA on DreamBooth, a subject-driven text-to-image personalization task that adapts a pretrained diffusion model from only a few reference images.
We provide side-by-side comparisons across methods to analyze identity preservation and prompt consistency across diverse prompts.

\subsection{Setup Details}
To evaluate LoCA on DreamBooth, we follow the experiment settings of \cite{hrnet} using Stable Diffusion v1.4. 
The benchmark comprises 30 classes with 25 prompts per class. 
Prompts follow the DreamBooth \cite{dreambooth} template ``a photo of [V] [C]'', where [V] is the identifier ``sks'' and [C] denotes the class name. 
Optimization utilizes the AdamW optimizer \cite{adamw} for 1,000 epochs. The learning rate is selected from {5e-5, 1e-4, 2e-4, 5e-4, 1e-3}, with 2e-4 as the default. 
For FSF \cite{chen2025large}, we report results using 1e-4, which exhibited superior performance in our comparative analysis.

\begin{figure}[t]
    \centering
    \setlength{\fboxsep}{0pt}
    \includegraphics[width=0.95\linewidth]{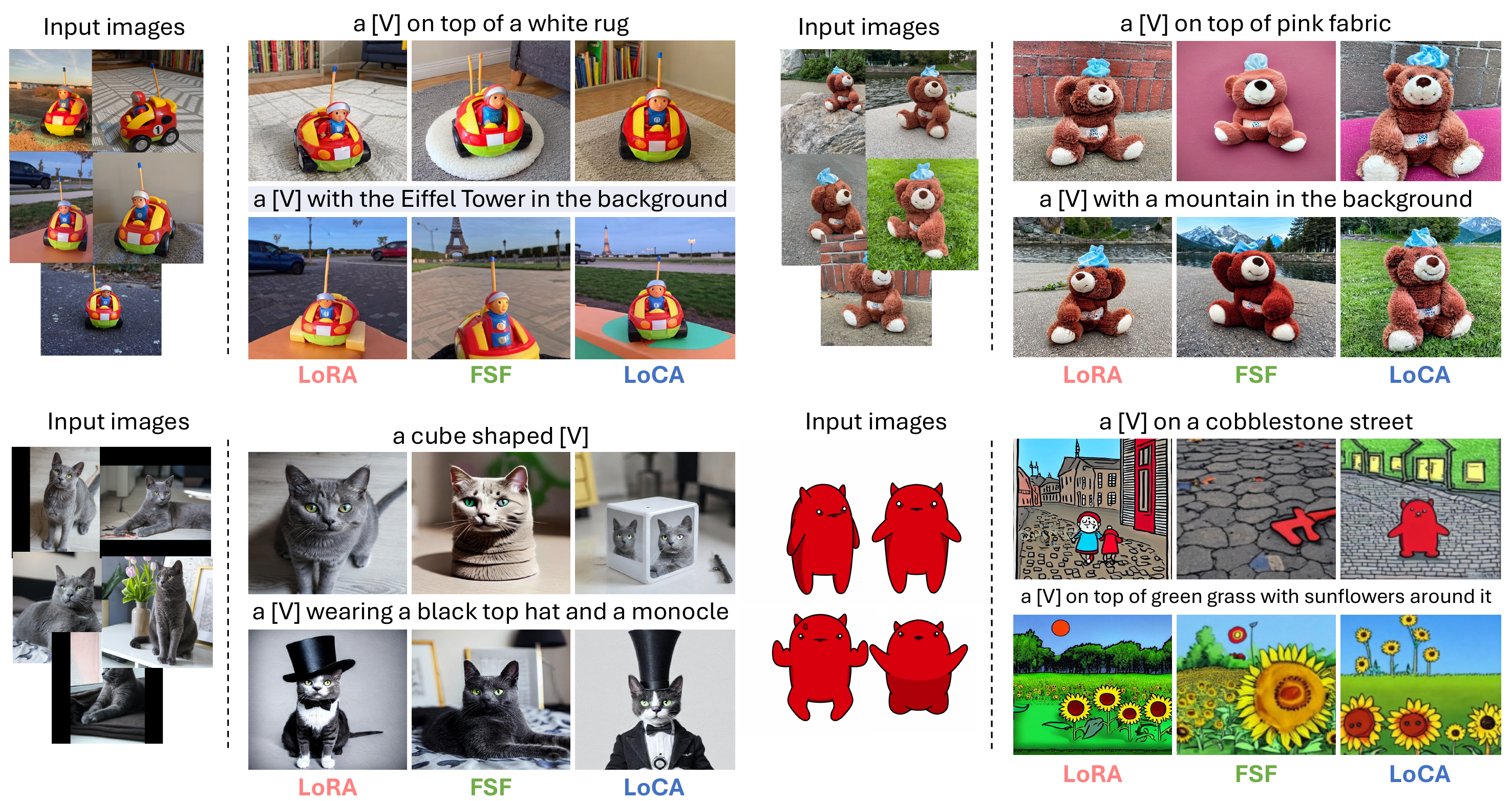}
    \caption{Qualitative results of subject-driven task. We visualize the results to compare PEFT methods: \colorbox{lorared}{\strut LoRA} \cite{lora}, \colorbox{sfgreen}{\strut FSF} \cite{chen2025large}, and \colorbox{oursblue}{\strut LoCA}.}
    \label{fig:dreambooth2}
    \centering
    \includegraphics[width=0.95\linewidth]{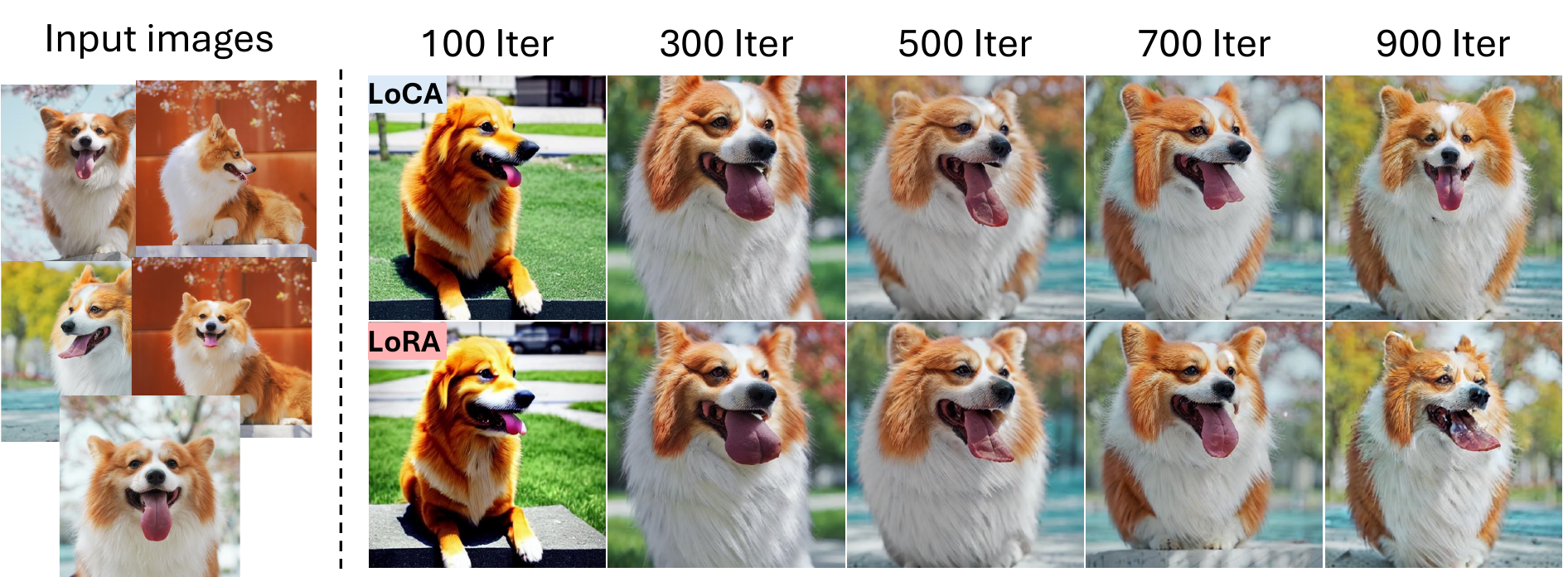}
    \caption{Qualitative results across training iterations. Both LoRA and LoCA show effective adaptation from around 300 iterations. LoRA shows a slight artifact around 900 iterations.}
    \label{fig:dreambooth_iter}
\end{figure}

\subsection{Experimental Results}
In \cref{fig:dreambooth2}, the difficulty of subject-driven generation depends heavily on both the prompt type and the input style. 
The top row presents relatively simple spatial and background-composition prompts, where all three methods successfully produce recognizable subject-centric images with marginal qualitative differences. 
Performance gaps emerge clearly when prompts demand explicit scene cues (e.g., the Eiffel Tower) rather than basic object placement.
The lower-left introduces the more complex challenges of geometry modification and attribute binding, accentuating the qualitative differences between methods. 
For the prompt ``a cube shaped [V],'' LoCA successfully captures the requested geometric transformation; the baselines incorrectly retain the original animal form. 
The prompt ``a [V] wearing a black top hat and a monocle'' further exposes varying degrees of success in rendering the specified attributes across the methods.
The lower-right panel poses the most challenging setting due to the non-photographic, cartoon-like input style. 
The models process the simpler prompt ``a [V] on a cobblestone street'' reasonably well, but struggle against the complex prompt ``a [V] on top of green grass with sunflowers.''
Overwhelmed by the detailed background request, all three methods over-emphasize the surrounding scene and noticeably degrade the original subject's identity.

\section{Domain Generalized Semantic Segmentation}
\label{sec:dg2}
We provide additional experimental details for the domain generalization benchmarks studied in Sec.~5.3, including domain generalized semantic segmentation (DGSS) and domain generalized object detection (DGOD).
These benchmarks assess adaptation to semantic segmentation and object detection under cross-domain distribution shift.

\subsection{Setup Details}
Our implementation uses the MMSegmentation \cite{mmseg2020} codebases for Domain Generalized Semantic Segmentation (DGSS) and Domain Generalized Object Detection (DGOD), respectively, and Hugging Face scripts for personalization experiments. For DGSS, we follow the configurations in SoMA \cite{yun2025soma}.
Mask2Former \cite{mask2former} serves as the default decode head using basic data augmentation from Rein \cite{rein}, and EMA is employed to ensure training stability. A consistent configuration is maintained by fixing both the rank and alpha at 16 across all experiments, including backbone adaptation. Optimization is performed using AdamW \cite{adamw} (lr = 0.001, weight decay = 0.05). Learning-rate multipliers of 0.5, 0.1, and 0.0 are applied to the backbone, $u_{dw}$, and $s_{basis}$, respectively. Notably, unlike the original SoMA, weight decay is disabled for these spatial SVD parameters ($u_{dw}$ and $s_{basis}$). We specifically freeze $s_{basis}$ with a multiplier of 0.0 to prevent basis collapse, as the adaptation relies primarily on the basis direction rather than its magnitude.

\begin{table}[t]
\caption{\small Effect of the proposed components under \textit{GTAV $\rightarrow$ Mapillary} DGSS setting. We highlight the best for each column.}
\vspace{-1em}
\centering
\small
\resizebox{\textwidth}{!}{
\begin{tabular}{l|c|ccccccccccccccccccc|c}
\toprule
Methods & Params. & road & side. & build. & wall & fence & pole & light & sign & vege. & terr. & sky & pers. & rider & car & truck & bus & train & motor. & bicy. & mIoU \\
\midrule
Full fine-tuning (baseline) & 87.56M & 
90.4 & 65.0 & 85.5 & 41.7 & 51.9 & 55.2 & 67.1 & 52.8 & 79.7 & \second{52.4} & 94.4 & \first{79.5} & \first{58.5} & \first{90.5} & \first{66.0} & 76.0 & 36.4 & 65.1 & 46.2 & 67.01 \\

$\llcorner$ + Channel Mixing & 5.0M &
\second{92.6} & \third{66.6} & \second{87.9} & \third{49.5} & \first{56.1} & \second{58.3} & \third{68.0} & \third{55.8} & \third{82.3} & \third{51.1} & \third{95.3} & \third{78.6} & 52.3 & \third{89.5} & \third{62.7} & \third{80.3} & \third{51.0} & \third{71.7} & \first{62.4} & \third{69.13} \\

$\llcorner$ + Spatial Basis & 5.0M &
\second{92.7} & \second{67.4} & \first{88.5} & \first{51.4} & \second{55.7} & \first{58.4} & \first{68.9} & \first{57.9} & \second{83.0} & 48.1 & \second{95.7} & \second{79.2} & \second{52.9} & \third{89.4} & 59.5 & \second{81.9} & \first{72.8} & \second{72.1} & \third{59.7} & \second{70.26} \\

$\llcorner$ + Hierarchical Rank & 5.0M &
\first{92.8} & \first{68.3} & \second{87.9} & \second{49.8} & \third{54.4} & \third{57.7} & \second{68.2} & \second{57.1} & \first{84.3} & \first{51.6} & \first{96.0} & \third{78.9} & \third{53.3} & \second{90.2} & \second{64.7} & \first{84.6} & \second{71.0} & \third{71.5} & \third{59.6} & \first{70.62} \\

\bottomrule
\end{tabular}}
\label{tab:ablation_components}
\end{table}

\subsection{Experimental Results}
In \cref{tab:ablation_components}, we report class-wise ablation results on the Mapillary dataset. Full fine-tuning achieves strong performance on several object-centric categories. Spatial basis modeling improves performance on classes with structured spatial patterns (e.g., \texttt{train} and \texttt{sign}). Hierarchical rank provides additional gains for categories that require broader spatial context, including \texttt{road} and \texttt{sidewalk}. These results indicate that different components contribute complementary benefits across category types.

\section{Evolution of Representational Capacity}
\label{sec:additional}

\begin{figure}[t]
    \centering
    \includegraphics[width=0.95\linewidth]{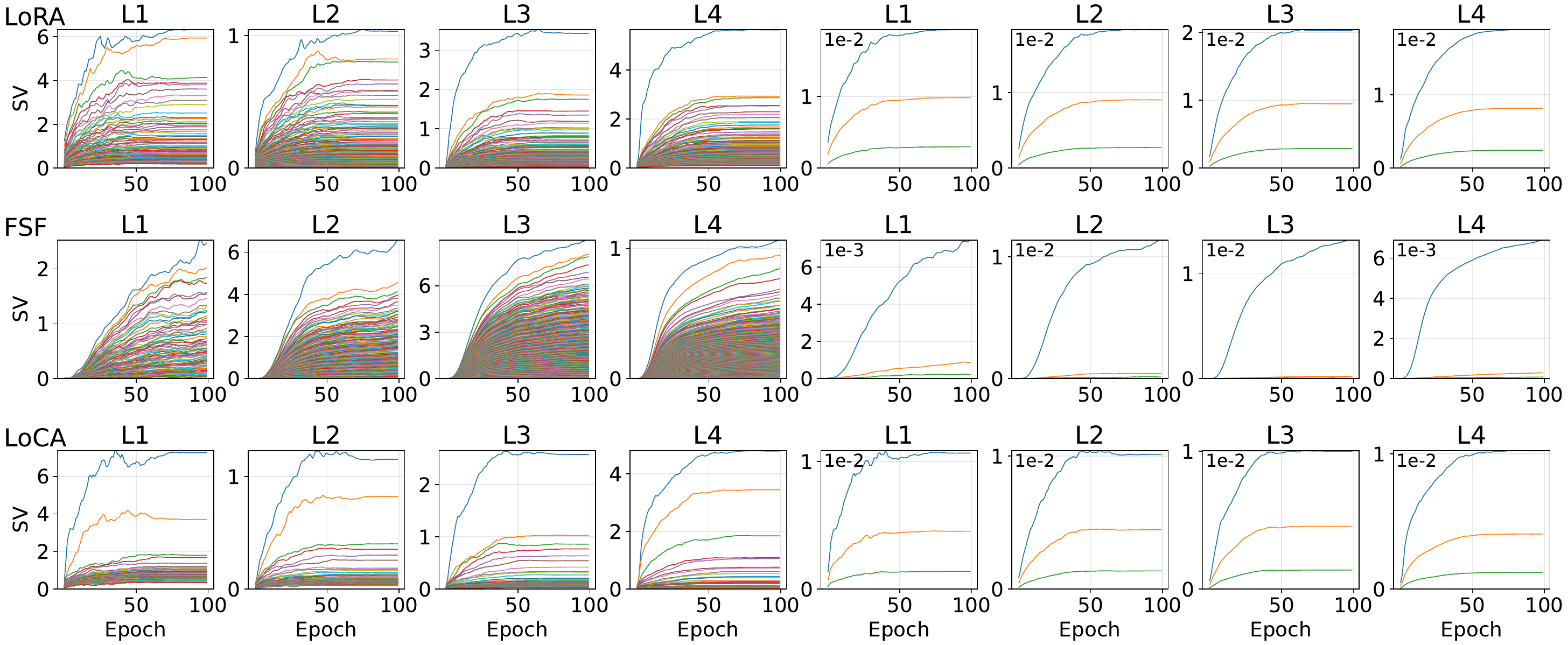}
    \vspace{-1em}
    \caption{Singular value evolution of ResNet-50 weights across PEFT methods (y-axis: singular values). Columns 1–4: channel-wise SVD of depthwise weights; columns 5–8: spatial SVD. L: layer.}
    \label{fig:sv_resnet}
    \includegraphics[width=0.95\linewidth]{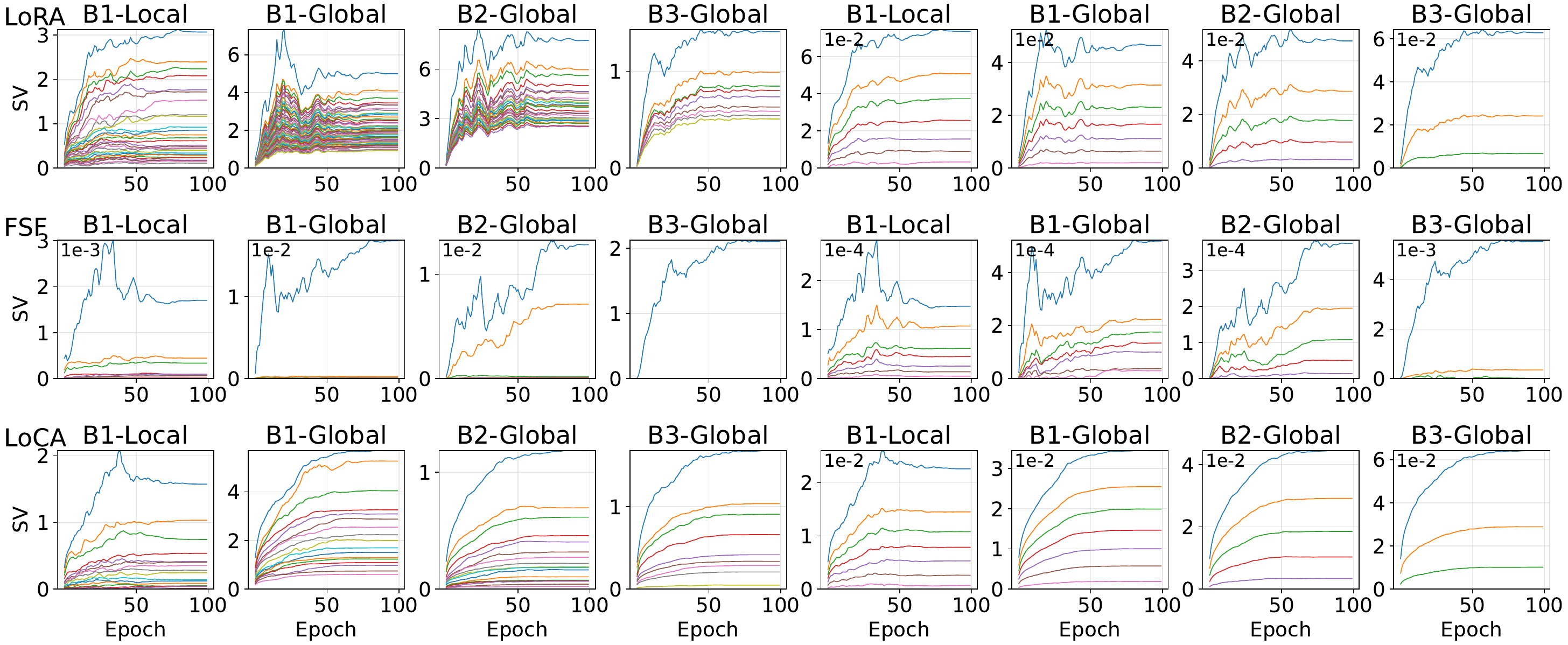}
    \vspace{-1em}
    \caption{Singular value evolution of MobileMamba-T2 weights across PEFT methods (y-axis: singular values). Columns 1–4: channel-wise SVD of depthwise weights; columns 5–8: spatial SVD. B: block. Local: depthwise convolution. Global: wavelet-domain convolution.}
    \label{fig:sv_mobilemamba}
\end{figure}

We visualize the Singular Value (SV) expansion results for ResNet-50 (\cref{fig:sv_resnet}) and MobileMamba (\cref{fig:sv_mobilemamba})  to complement the analysis presented in the main paper. The SV distributions across layers provide a qualitative view of adaptation behavior.

\begin{figure}[t]
    \centering
    \includegraphics[width=\linewidth]{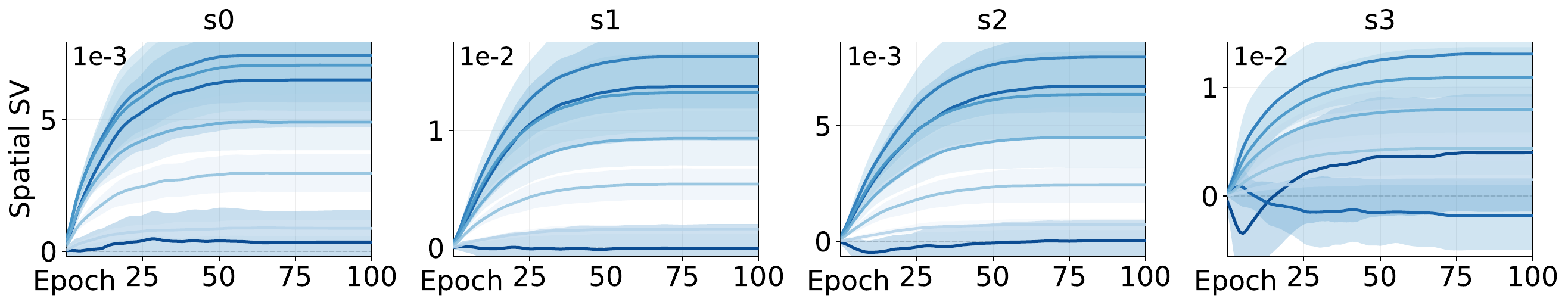}
    \vspace{-8mm}
    \caption{SV expansion of ConvNeXt-B on VTAB-1K. Shaded regions denote standard deviation across datasets. \textbf{LoCA shows consistent adaptation expressivity across datasets.}}
    \label{fig:spatial_basis_expansion}
\end{figure}

\noindent\textbf{ResNet.}
For ResNet-50 (\cref{fig:sv_resnet}), FSF exhibits slow growth during the early training stage. The SVs gradually expand as training progresses. This trend indicates that adaptation occurs but proceeds inefficiently. In addition, the spatial components show limited SV expansion. LoRA and LoCA show faster SV expansion in the ResNet architecture. LoRA produces stronger SV expansion in the early layers. LoCA shows a more consistent SV distribution across all layers. LoRA tends to concentrate adaptation on early features. LoCA reflects features more uniformly across the network. The stacked convolutional structure of ResNet is associated with relatively moderate SV expansion compared to other architectures.

\noindent\textbf{MobileMamba.}
In the case of MobileMamba-T2 (\cref{fig:sv_mobilemamba}), the architecture employs $3\times3$ convolution kernels for local learning and wavelet-domain convolutions to capture spatial global information in each block. The wavelet kernels are defined with different sizes (3, 5, and 7) across blocks. This hierarchical design facilitates multi-scale representation learning. LoRA exhibits rapid SV expansion during early training. The SV values in the later components fluctuate noticeably. The degree of SV expansion varies across blocks. FSF also exhibits fluctuations and inconsistent SV expansion between blocks.  LoCA shows a different pattern. Some fluctuations appear in the local components. The global components remain stable. The SV values adjust smoothly across kernels. Both channel and spatial information are updated consistently.

\section{Additional Sensitivity Analyses}
\label{sec:additional_results}
\begin{figure}[t]
    \centering
    \begin{subfigure}{0.48\columnwidth}
        \centering
        \includegraphics[width=\linewidth]{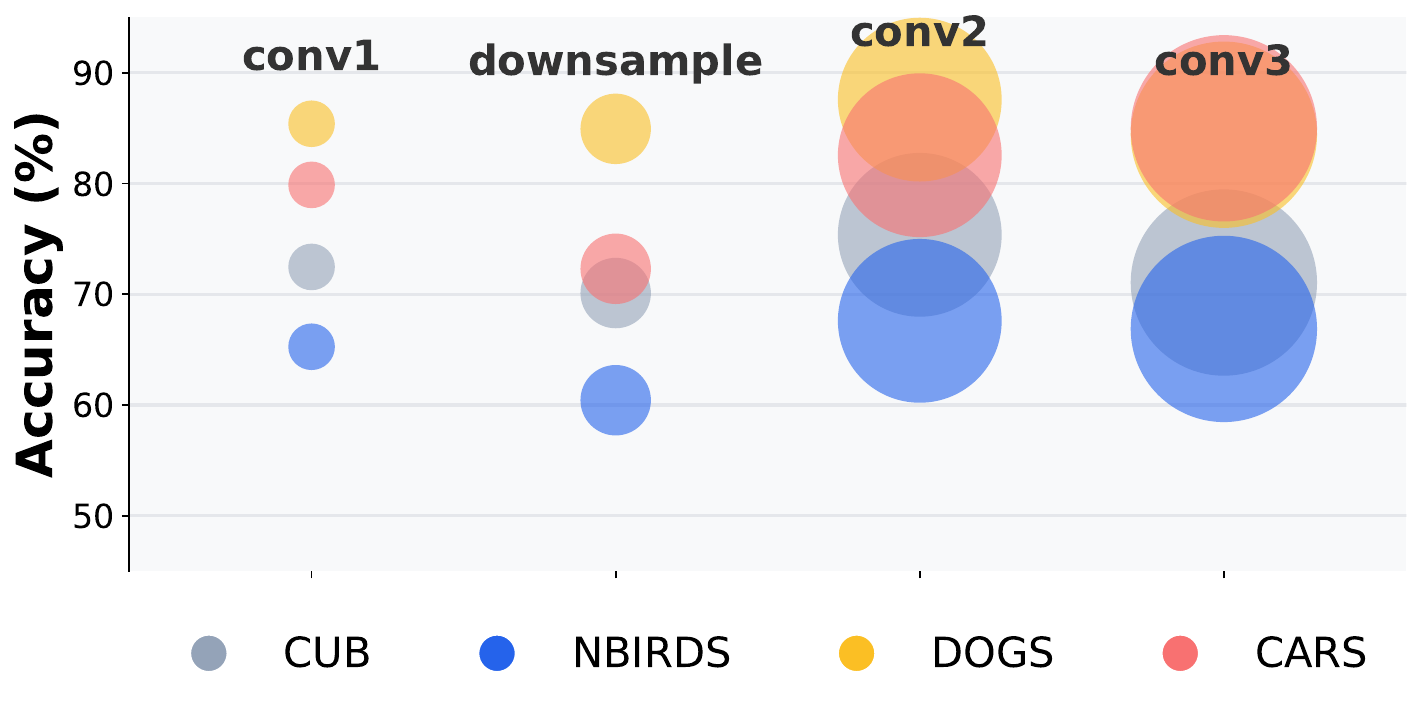}
        \caption{ResNet-50 performance}
        \label{fig:target_resnet}
    \end{subfigure}
    \hfill
    \begin{subfigure}{0.48\columnwidth}
        \centering
        \includegraphics[width=\linewidth]{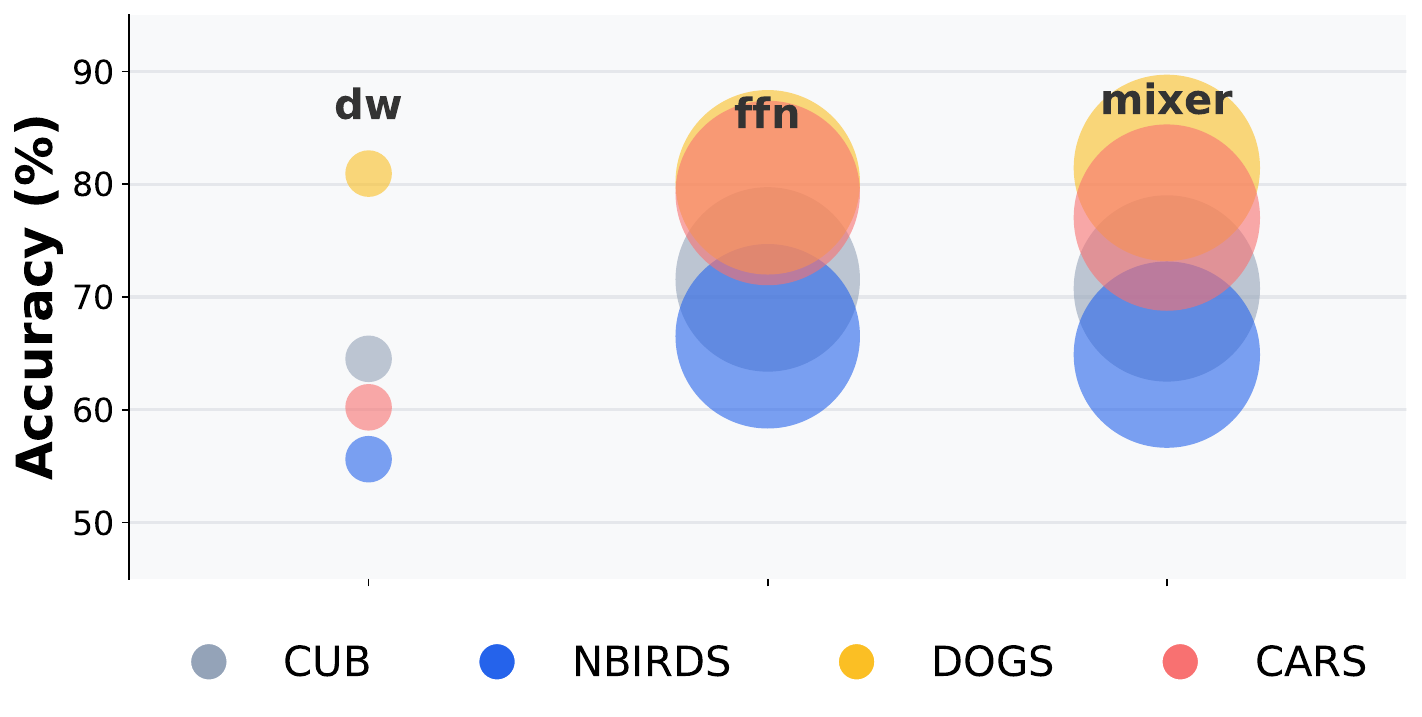}
        \caption{MobileMamba-T2 performance}
        \label{fig:plot2}
    \end{subfigure}
    \vspace{-1em}
    \caption{Performance across target modules on FGVC datasets. Each bubble's area is proportional to the number of parameters.}
    \label{fig:modules}
\end{figure}
\subsection{How sensitive is adaptation performance to the choice of convolution modules with different kernel structures?}
To analyze performance gains across target modules, ResNet-50 and MobileMamba-T2 are evaluated with all ranks set to 16. ResNet consists of 1×1 and 3×3 convolutions with downsample layers. The conv1 and conv2 modules correspond to the two 3×3 convolutions, while conv3 corresponds to the 1×1 convolution. \cref{fig:modules} shows that adapting only
conv3 achieves performance comparable to other configurations while requiring fewer parameters. MobileMamba adopts a hybrid architecture combining convolution and linear operations. The modules are divided into three components: depthwise convolution (dw), feed-forward network (ffn), and mixer. The dw captures local spatial patterns through channel-wise filtering, the ffn performs channel mixing through pointwise convolutions, and the mixer models long-range dependencies using an SSM-based operator. Adapting the mixer yields the largest performance improvement, whereas adapting only dw captures mainly local information and exhibits performance variance. We observe that the adaptation performs consistently across convolution modules, indicating stable behavior across layers.

\subsection{Sensitivity Analysis of the Scaling Factor}
\begin{figure}[h]
    \centering
    \includegraphics[width=\linewidth]{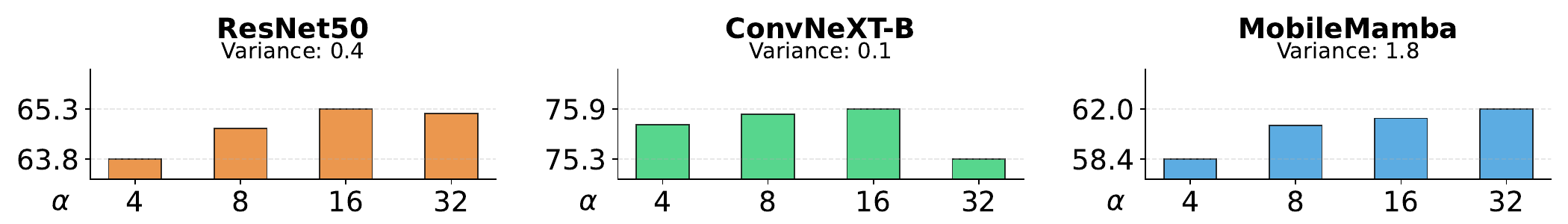}
    \vspace{-2em}
    \caption{Sensitivity analysis of the scaling factor $\alpha$}
    \label{fig:alpha_comparison}
\end{figure}

Since LoRA has been reported to be sensitive to the scaling factor $\alpha$~\cite{biderman2024lora}, we evaluate $\alpha \in {4, 8, 16, 32}$ across three architectures.
As shown in \cref{fig:alpha_comparison}, LoCA maintains consistent accuracy across ResNet-50, ConvNeXt-B, and MobileMamba-T2, indicating that LoCA is robust to the choice of $\alpha$.
\vspace{-4mm}

\section{Code}
\label{sec:code}

In \cref{sec:code}, we provide a reference implementation for the LoRA and LoCA formulations discussed in Sec. 4.
Code~\ref{code:lora}\footnote{We use the official implementation of LoRA available at 
\url{https://github.com/microsoft/LoRA/blob/main/loralib/layers.py}. 
All LoRA experiments are conducted using this implementation.} shows the flattened convolutional LoRA baseline.
Code~\ref{code:loca} shows how LoCA composes a channel low-rank update with SVD-initialized spatial basis refinement on top of a pretrained convolution kernel.

\lstset{
    language=Python,
    basicstyle=\ttfamily\footnotesize,
    breaklines=true,
    frame=single,
    mathescape=true
}
\renewcommand{\lstlistingname}{Code}
\begin{lstlisting}[caption={Baseline convolutional LoRA implementation}, label={code:lora}]
class ConvLoRA(nn.Module):
    def __init__(self, conv_module, in_channels, out_channels,
                 kernel_size, r=0, lora_alpha=1):

        super().__init__()
        self.conv = conv_module(in_channels, out_channels, kernel_size)

        # LoRA performs low-rank decomposition on the flattened convolution weight
        # Assume square kernel (kh = kw)
        # lora_a: [r * kh, in_channels * kh]
        # lora_b: [out_channels * kh, r * kh]
        self.lora_a = nn.Parameter(
            torch.zeros(r * kernel_size, in_channels * kernel_size)
        )
        self.lora_b = nn.Parameter(
            torch.zeros(out_channels * kernel_size, r * kernel_size)
        )
        self.scaling = lora_alpha / r

    def forward(self, x):
        delta_w = (self.lora_b @ self.lora_a)
        delta_w = delta_w.view(self.conv.weight.shape)

        return self.conv._conv_forward(
            x,
            self.conv.weight + delta_w * self.scaling,
            self.conv.bias
        )
        return self.conv(x)
\end{lstlisting}

\begin{lstlisting}[caption={Proposed LoCA implementation}, label={code:loca}]
class LoCAConv2d(nn.Module):
    def __init__(self, conv, lora_rank=16, lora_alpha=16.0):
        super().__init__()

        self.weight_shape = conv.weight.shape
        out_c, in_c, kh, kw = self.weight_shape

        self.weight = nn.Parameter(conv.weight.data.clone(), requires_grad=False)
        self.stride = conv.stride
        self.padding = conv.padding
        self.dilation = conv.dilation
        self.groups = conv.groups

        self.r = min(lora_rank, out_c)
        self.scaling = (lora_alpha / self.r)

        # Sec. 4.1 Low-Rank Channel Adaptation
        # Channel mixing low-rank update for convolution kernel
        # lora_a: [r, in_channels * kh * kw]
        # lora_b: [out_channels, r]
        self.lora_a = nn.Parameter(torch.empty(self.r, in_c * kh * kw))
        self.lora_b = nn.Parameter(torch.zeros(out_c, self.r))

        # Spatial basis extracted from pretrained conv kernels
        # s_basis: [kh*kw, kh, kw]
        self.spatial_rank = kh * kw
        self.s_basis = nn.Parameter(init_spatial_basis_from_svd(conv.weight))
        
        # Spatial update coefficients for diagonal channel pairs
        # u_dw: [diag (min(out_c,in_c)), spatial_rank]
        self.diag = min(out_c, in_c)
        self.u_dw = nn.Parameter(torch.zeros(self.diag, self.spatial_rank))

    # Sec. 4.2 SVD-based Spatial Basis Refinement
    # Spatial diagonal update
    def _get_delta_weight(self):
        out_c, in_c, kh, kw = self.weight_shape

        dW_ch = (self.lora_b @ self.lora_a) * self.scaling

        dW_sp_diag = self.u_dw @ self.s_basis.view(self.spatial_rank, -1)

        dW = dW_ch.view(out_c, in_c, kh * kw)
        dW.diagonal(dim1=0, dim2=1).add_(dW_sp_diag.T)

        return dW.reshape(out_c, in_c, kh, kw)

    def forward(self, x):
        W_eff = self.weight + self._get_delta_weight()

        return F.conv2d(
            x, W_eff, self.bias,
            stride=self.stride,
            padding=self.padding,
            dilation=self.dilation,
            groups=self.groups
        )
\end{lstlisting}

\section{Limitation and Future Work}
\label{sec:limit}

The proposed framework implements a PEFT approach for convolutional operators by decoupling channel mixing and spatial basis components. This formulation facilitates efficient adaptation and improves the parameter–performance trade-off compared to LoRA. The current limitation stems from the fixed spatial basis design across varying kernel sizes, which constrains the spatial representation capacity of 1x1 kernels. Nevertheless, empirical evaluations demonstrate robust performance across diverse kernel scales. Future research aims to investigate adaptive construction of spatial bases conditioned on specific kernel dimensions.

\clearpage

\end{document}